\pdfoutput=1

\documentclass[11pt]{article}

\usepackage[]{EMNLP2022}

\usepackage{times}
\usepackage{latexsym}
\usepackage{booktabs}

\usepackage[T1]{fontenc}

\usepackage[utf8]{inputenc}

\usepackage{microtype}

\usepackage{inconsolata}

\usepackage{graphicx,subfigure}
\usepackage{amsmath}
\usepackage{amssymb}
\usepackage{mathtools}
\usepackage{amsthm}
\usepackage{comment}
\usepackage{algorithm}
\usepackage{algorithmic}
\usepackage{bbold}
\definecolor{bettergreen}{rgb}{0.13, 0.55, 0.13}

\newcommand{\editms}[1]{}
\renewcommand{\editms}[1]{{{#1}}}

\newcommand{\modelname}[1]{}
\renewcommand{\modelname}[1]{{\color{black} \textsc{Referee}}}

\newcommand{\firstphasemodelname}[1]{}
\renewcommand{\firstphasemodelname}[1]{{\color{black} \textsc{Referee-Distill}}}
\newcommand{\secondphasemodelname}[1]{}
\renewcommand{\secondphasemodelname}[1]{{\color{black} \textsc{Referee-Control}}}

\usepackage[capitalize,noabbrev]{cleveref}

\title{
\modelname{}: Reference-Free Sentence Summarization\\ with Sharper Controllability 
through Symbolic Knowledge Distillation}

\author{Melanie Sclar\textsuperscript{1} \ \ \ \ Peter West\textsuperscript{1} \ \ \ \ Sachin Kumar\textsuperscript{2} \ \ \ \ Yulia Tsvetkov\textsuperscript{1} \ \ \ \ Yejin Choi\textsuperscript{1,3} \\
\textsuperscript{1}Paul G. Allen School of Computer Science \& Engineering, University of Washington \\
\textsuperscript{2}Language Technologies Institute, Carnegie Mellon University\\
\textsuperscript{3}Allen Institute for Artificial Intelligence \\
\texttt{msclar@cs.washington.edu}
}

\begin{document}
\maketitle

\begin{abstract}
We present \modelname{}, a novel framework for sentence summarization that can be trained reference-free (i.e., requiring no gold summaries for supervision), while allowing direct control for compression ratio.  
Our work is the first to demonstrate that reference-free, controlled sentence summarization is feasible via the conceptual framework of Symbolic Knowledge Distillation \citep{west2021symbolic}, where latent knowledge in pre-trained language models is distilled via explicit examples sampled from the teacher models, further purified with three types of filters: length, fidelity, and Information Bottleneck. 
Moreover, we uniquely propose \emph{iterative} distillation of knowledge, where student models from the previous iteration of distillation serve as teacher models in the next iteration. 
Starting off from a relatively modest set of GPT3-generated summaries, we demonstrate how iterative knowledge distillation can lead to considerably smaller, but better summarizers with sharper controllability. 
A useful by-product of this iterative distillation process is a high-quality dataset of sentence-summary pairs with varying degrees of compression ratios. 
Empirical results demonstrate that the final student models vastly outperform the much larger GPT3-Instruct model in terms of the controllability of compression ratios, without compromising the quality of resulting summarization.\footnote{\editms{See \url{https://github.com/msclar/referee} for code, models, and data.}} 

\end{abstract}

\section{Introduction}

%

We introduce \modelname{}, a new framework for sentence summarization that works by iteratively generating and distilling knowledge into successively better models. This allows \modelname{} to be [Refer]ence fr[ee]---beginning by distilling from a large language model rather than with supervised data. Yet, our method results in a more efficient, compact, and controllable summarization model than what we start with.

\definecolor{mycyan}{RGB}{22,231,207}
\definecolor{mygreen}{RGB}{97,216,54}
\definecolor{myblue}{RGB}{0,162,255}
\definecolor{mygray}{RGB}{200,200,200}
\definecolor{mygray2}{RGB}{214,213,214}
\definecolor{myblack}{RGB}{0,0,0}

\begin{figure}[t]
    \centering
    \includegraphics[width=0.98\linewidth]{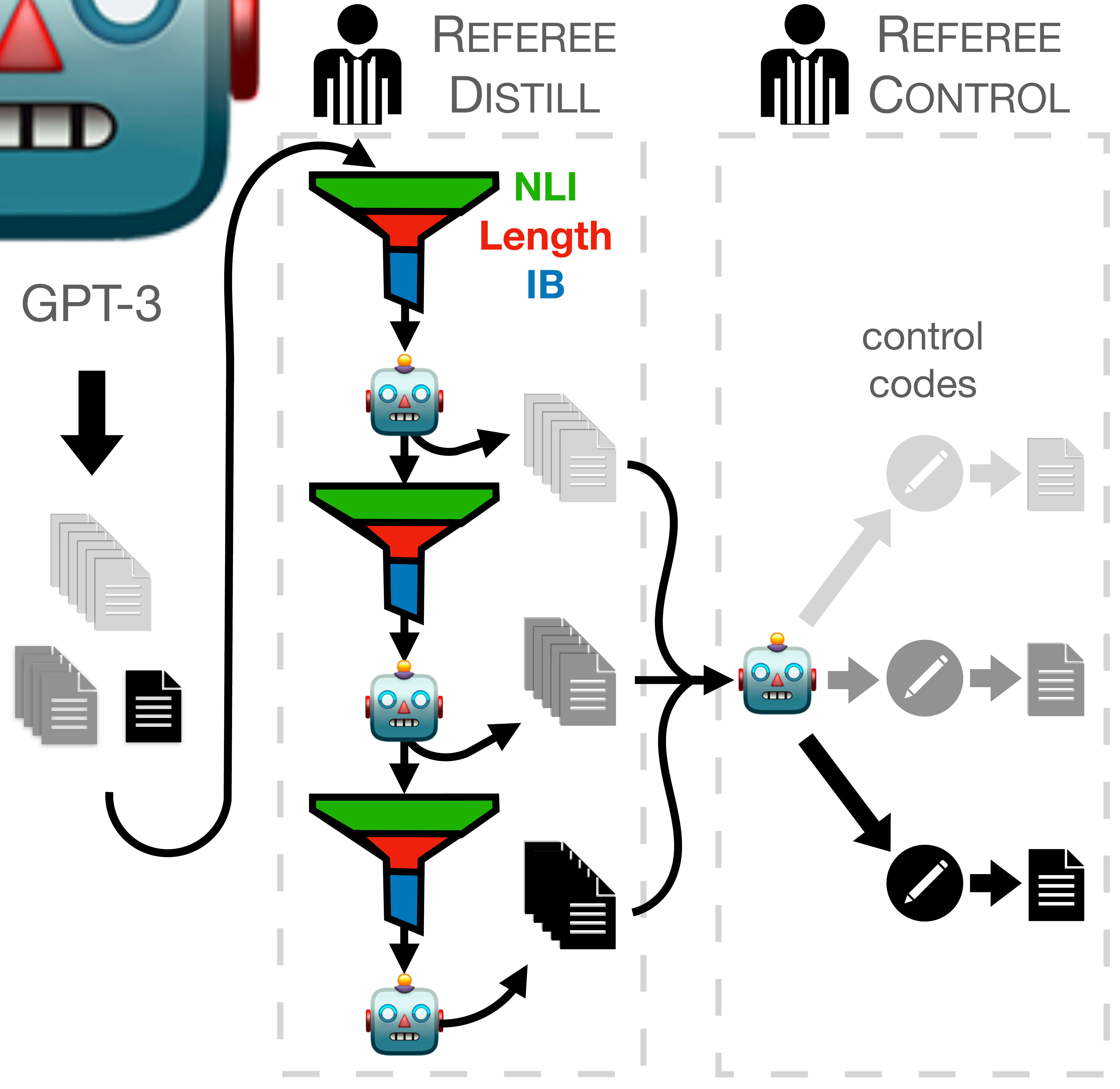}
    \caption{Our method results in high quality, \emph{reference-free} compact summarizers. We begin by using a large language model (e.g. GPT-3) to generate many summaries that demonstrate different aspects we may want in a summary---\textbf{\color{mygray} gray} represents an aspect well-represented in these generations, while \textbf{\color{myblack} black} is underrepresented. We first use \firstphasemodelname{} to iteratively filter and train summarizers that better represent these desirable aspects, e.g. shorter summary length. We then use generations from \firstphasemodelname{} to train a model in which these aspects are controllable: this is \secondphasemodelname{}.
    }
    \label{fig:1}
\end{figure}

Our work follows the paradigm of Symbolic Knowledge Distillation \cite{west2021symbolic}, which transfers implicit knowledge from a massive language model to a considerably smaller student model by explicitly generating knowledge in textual form. Unlike traditional knowledge distillation \citep{hinton2015distilling} where the teacher model and the student model are of the same type, symbolic knowledge distillation allows for the student model to be of a different type. 

Our work differs from~\citet{west2021symbolic} in three key aspects. First, our distillation is \emph{iterative}: each student model becomes a teacher in successive rounds, refining and improving summarization at every step. Second, \modelname{} controls for more than just overall quality, improving multiple model aspects in each round such as length, fidelity, and information bottleneck~\citep{tishby99information}, then allowing explicit length control at generation time. Third, our work is the first 
to {show} 
that reference-free, controlled sentence summarization can be formulated as symbolic knowledge distillation. 

\modelname{} works in two phases, illustrated in Figure~\ref{fig:1}. First, \firstphasemodelname{} uses a modest number of generated summaries from GPT-3 \cite{brown2020language} to produce high quality and compact summarizers \citep{goyal2022news}. We follow an iterative approach; in each iteration we filter generations for desirable qualities, re-train a new and better summarizer, and finally generate new summaries for the next round. Each round amplifies effects of the previous rounds, improving notions of summary quality like entailment or shorter length. Second, \secondphasemodelname{} uses these iteratively distilled summaries to train a model with explicit control: in our experiments, we use progressively {shortened} generations from each iteration to train a final summarizer with explicit length control. 


We find that \modelname{} demonstrates compelling empirical results compared to competitive baselines. \firstphasemodelname{}, even without explicit length control, is able to generate shorter summaries with more consistency and equal quality compared with the original teacher model (GPT-3, 16x larger in size)  as well as a supervised model. Moreover,  \secondphasemodelname{}, which has more direct length control baked in, demonstrates a sharp  degree of control in length, and succeeds at generating high quality summaries at specified lengths with significantly higher accuracy than GPT-3. 
%
In sum, the promising empirical results of \modelname{} encourages further future investigation to extend the framework of symbolic knowledge distillation for reference-free, controlled text summarization.

\section{Methods} 

We first describe \firstphasemodelname{} (see \S\ref{sec:iterative_distill}), an iterative procedure to promote specific behaviors that may not be prevalent in the original data, while maintaining summary quality. We explore two different filters, detailed in \S\ref{sec:filters}.
We then detail \secondphasemodelname{} (see \S\ref{sec:iterative_control}), a model that separates summaries into categorical variables and is iteratively trained to, summarize a given sentence within the desired 
category (e.g., a range of compression ratio). In this work we only consider  
categories that reflect different compression ratio, but the same approach could be applied to other types of control categories, such as style.

\subsection{Iterative Symbolic Knowledge Distillation: \firstphasemodelname{}}\label{sec:iterative_distill}

Let $\mathcal{D} = \mathcal{D}_0{ }\cup \ldots \cup{ }  \mathcal{D}_t$ denote a sentence corpus without reference summaries. 
We start with a teacher model (GPT3-Instruct Curie) from which we want to distill summarization knowledge under a fixed budget. Using $\mathcal{D}_0$---a small subset of $\mathcal{D}$---we first generate a dataset of sentence-summary pairs ($\mathcal{C}_0)$ by few-shot prompting the teacher {and automatically filtering low-quality generations}. \editms{Filters will be detailed in Section \ref{sec:filters}}. Throughout the whole training procedure, we store each entry $(s, s')$ as ``\texttt{$s$ TL;DR: $s'$ <eos>}''. Here, \texttt{<eos>} denotes end of sequence and \texttt{TL;DR;} is a separator that has been shown to encourage summarization behavior \citep{radford2019language}.


Let $\mathcal{M}_0$ be a pre-trained model significantly smaller than GPT-3 (GPT2-Large in our experiments). Using the seed dataset $\mathcal{C}_0$, we train a student model $\mathcal{M}_1$ by fine-tuning $\mathcal{M}_0$ with language modeling loss. 
We then iteratively refine this model by (1) using it to generate summaries for a subset of $\mathcal{D}$, (2) filtering them to remove undesired behaviors, and (3) training another student model on the filtered dataset, essentially distilling a better summarizer. More precisely, 
\begin{align*}
    \mathcal{C}_{i} &:= \mathrm{filter}_i(\mathrm{generate}(\mathcal{M}_{i}, \mathcal{D}_{i}))\\
\mathcal{M}_{i+1} &:= \mathrm{finetune}(\mathcal{M}_i, \mathcal{C}_i)
\end{align*}

\noindent
We execute this procedure for $t$ steps, creating $t\!+\!1$ different summarization datasets in the process: $\mathcal{C}_0, \mathcal{C}_1, \ldots, \mathcal{C}_t$.\footnote{Note that this process would stay identical if a user decided to use a human-generated summarization dataset as $\mathcal{C}_0$.}
We discuss two possible instantiations of the $\mathrm{filter}_i$ below. 

\subsection{Filters}\label{sec:filters}
There is no one summary that is better than all others; depending on the desiderata of the end users, some might prefer shorter but less informative summaries, while others might prefer longer, and more informative ones.
While some of these goals are universal and always desired (for example, a summary should be accurate, in that it should not contain information not present in the input), others can be tailored to the end task. We use {binary filters} ($\mathrm{filter}_i$) to operationalize these goals. We experiment with the following filters. 



\paragraph{Summary Fidelity Filter} 
To encourage accurate summaries, we employ a simple but effective criterion: the summary should be entailed by the input sentence. More formally, we define a binary filter, $f_\textsc{NLI}(s, s'):=\mathbb{1}\{s \Rightarrow s'\}$, and discard all non-entailed sentence-summary pairs to avoid using these samples when training the next iteration's student. We measure entailment using an off-the-shelf state-of-the-art NLI model~\citep{liu2022wanli}.

\paragraph{Summary Length Filter}
While underexplored in prior work, constraining for the length of written text, especially in summarization, is a desirable feature to support real world applications with limited screen space.  
To obtain a corpus of summaries of varying lengths, at each distillation step $i$, we encourage the student $\mathcal{M}_i$ to generate progressively shorter outputs. We achieve this by constraining $\mathcal{C}_i$ to contain only summaries with a predefined compression ratio $r_i \in [0, 1]$. More precisely,
\begin{align*}
    f_\textrm{compress}(s, s', r_i) &= \mathbb{1}\Big\{\frac{|s'|}{|s|} \leq r_i\Big\}
\end{align*}
where $r_i > r_{i+1}$ for all $i$, to progressively summarize more succinctly. $\frac{|s'|}{|s|}$ is commonly referred to as \textit{compression ratio}.
In theory, one could generate data for all desired compression ratios directly from $\mathcal{M}_1$. However, since the seed dataset $\mathcal{C}_0$ is heavily skewed towards longer summaries, the final corpus after filtering with $f_\mathrm{NLI}$ would be extremely small for lower compression ratios. We find that combining the two filters and iteratively refining models to produce shorter, accurate summaries leads to a more diverse and still high-quality final corpus.
\paragraph{Contextual Filter}
For many applications, the sentences we need to summarize are part of a larger piece of text, such as a paragraph or a document (e.g. emails, articles). 
This contextual information may further improve sentence summary quality, since depending on the larger context, different information could be more important to be preserved, and inter-sentence redundancies could be removed. 
Inspired from \citet{west-etal-2019-bottlesum}'s interpretation of the Information Bottleneck principle~\citep{tishby99information}, we consider the following filter:
\begin{align*}
f_\mathrm{NSP} = \mathbb{1}\Big\{\frac{p(s_\mathrm{next}|s')}{p(s_\mathrm{next}|s)} \geq l\Big\}
\end{align*}
where NSP refers to ``next sentence prediction'', $p$ is an oracle language model (which we approximate by GPT2-Large), $s_\mathrm{next}$ denotes the sentence immediately following the input sentence $s$, and $l \in [0, 1]$ is a hyperparameter. 
Intuitively, we want to find summaries which are good predictors of the next sentence, to select the most crucial information and preserve coherence. $l$ allows us to strike a balance between sacrificing some of the information in $s$ and maintaining enough to predict $s_\mathrm{next}$.


Adding $f_\mathrm{NSP}$ requires expanding the input sequence to also include the next sentence throughout the iterative distillation process defined in \S \ref{sec:iterative_distill}.

\paragraph{Final \firstphasemodelname{} Filters Definition}
We experiment with two filters, $f_1$ and $f_2$ {(or \#1 and \#2, as we will refer to during experiments)}. $f_1$ does not assume the existence of any context, and so it only filters for inaccuracies and length:
\begin{align*}
f_1(s, s'; s_\mathrm{next}, r_i) = f_\mathrm{NLI} \wedge f_\mathrm{compress}
\end{align*}
This allows $f_1$ to be applied in broader contexts. We also define $f_2$, which adds contextual filtering:
\begin{align*}
f_2(s, s';  s_\mathrm{next}, r_i, l) = f_\mathrm{NLI} \wedge f_\mathrm{compress} \wedge f_\mathrm{NSP}
\end{align*}

\paragraph{Fluency Filter} To ensure fluency over several self-training iterations, we consider an additional filter only to be used in \secondphasemodelname{}. 
Given a sentence $\mathbf{x} = (x_1,\ldots, x_\ell)$, 
we define $\mathrm{AvgNLL}(\mathbf{x}) := -\frac{1}{\ell}\sum_{i\leq \ell}{ \log{p(x_i|x_{<i})}}$. 
We determine a summary as fluent if and only if its mean \editms{Negative Log Likelihood (NLL)} does not exceed that of source sentence, 
leading to the filter:
\begin{align*}
    f_\textrm{AvgNLL}(s, s') &= \mathbb{1}\Big\{\textrm{AvgNLL}(s') \leq \textrm{AvgNLL}(s)\Big\}
\end{align*}

\subsection{\secondphasemodelname{}}\label{sec:iterative_control}


Using the high quality corpora of varying compression ratios obtained using \firstphasemodelname{}, we train
\secondphasemodelname{}, a summarization model that allows explicit control for desired compression ratio. 
We divide all possible compression ratios into $n$ buckets, where each bucket 
$b_i\!=\!\big[\frac{i}{n},\!\frac{i+1}{n}\big]$ for $0\!\leq\!i\!<\!n$.
Using $b_i$ as control codes, we train a model that, when prompted with it, can summarize at a compression ratio within $b_i$.

Similar to $\mathcal{D}$, we start with a corpus $\mathcal{F} = \mathcal{F}_0 \cup \ldots \cup \mathcal{F}_t$ of sentences without reference summaries.
Additionally, we create a seed corpus labeled with compression ratios, $\mathcal{E}_0 = \mathcal{C}_0\cup \ldots \cup \mathcal{C}_t$ {($\mathcal{F}_0 = \mathcal{D}_0\cup \ldots \cup \mathcal{D}_t$)} now representing each example $(s, s')$ as ``\texttt{$s$ <sep> <bucket\_tok $j$> TL;DR: $s'$ <eos>}'', where \texttt{<bucket\_tok $j$>} corresponds to the bucket in which the example lies, that is $\frac{|s'|}{|s|} \in b_j$. \texttt{<sep>} is a special token. 
We denote each subset of $\mathcal{E}_0$ corresponding to bucket $j$ as $\mathcal{E}_0^{(j)}$. {This seed dataset is filtered to remove low-quality generations, with the same filter as all the subsequent iterations.}

Similar to \firstphasemodelname{}, starting with a pre-trained model $\mathcal{N}_0$ (GPT2-Large), we train student models via iterative distillation. In each iteration $i$, (1) we fine-tune the student model using the bucket labeled corpus $\mathcal{E}_i$, (2) generate summaries for $\mathcal{F}_i$ for all buckets, (3) filter them to create a new labeled corpus $\mathcal{E}_{i+1}$.  
{We do not reinitialize the student at each iteration, but rather fine-tune starting from the teacher's current local optima.} 
We use $h(s, s') = f_{\mathrm{NLI}} \wedge f_{\mathrm{AvgNLL}}$ as the filter. Formally,
\begin{align*}
    \mathcal{N}_{i+1} &:= \mathrm{finetune}(\mathcal{N}_i, {\mathcal{E}^{(0)}_{i}, \mathcal{E}^{(1)}_{i}, \ldots, \mathcal{E}^{(n-1)}_{i}})\\
    \mathcal{E}^{(j)}_{i+1} &:= h(\mathrm{generate}(\mathcal{N}_{i}, \mathcal{F}_{i}, j)) \text{ \ $\forall \ 0\!\leq\!j\!<\!n$}
\end{align*}


\subsection{Primal-Dual Problem Interpretation of Summarization}\label{sec:dual}

Assuming summaries are fluent and factual, sentence summaries trade off between two variables: level of compression and level of information preservation. We are able to effectively fix the level of compression by introducing control codes, and then develop models to maximize information preservation. This is our primal problem. Thanks to length-control codes, we can now also solve the dual problem: ``what is the best shortest summary we could write?''. Written more precisely, given a fixed level of tolerance for losing information from the original sentence, what is the shortest summary we could write?
Furthermore, comparing similar-lengthed summaries also allows for fairer comparisons, since we are effectively measuring changes in only one variable.

\section{On GPT3's Fidelity and Length Control}\label{sec:analyze-gpt3}


We analyze GPT3-Instruct Curie's \citep{brown2020language} sentence summarization capabilities. 
We promote GPT3 to summarize at different compression ratios by few-shot prompting with high-quality sentence-summary pairs in the desired compression ratios. More precisely, we do three-shot prompting with three different sets of summaries: one set of sentence-summary pairs has all three pairs with compression ratios in the interval $[0.6, 0.8]$, another set in $[0.4, 0.6]$, and another in $[0.2, 0.4]$.

We show that average compression ratio (c.r.) correlates with the prompts' compression ratio (although variance is large), and up to 33\% of the time models generate summaries longer than the original sentence (see Table \ref{table:statistics-iterative})
. Qualitatively, this seems to be because of punctuation edits or hallucinations. 

\begin{table}[]
\footnotesize

\centering
\begin{tabular}{@{}ccccc@{}}
\toprule
\multicolumn{1}{c}{\textit{\begin{tabular}[c]{@{}c@{}}Model \end{tabular}}} & \textit{\begin{tabular}[c]{@{}c@{}}Avg. \\ c.r. \%\\ (stdev)\end{tabular}} & \textit{\begin{tabular}[c]{@{}c@{}}WANLI \\ Entailment\\ \%\end{tabular}} & \textit{\begin{tabular}[c]{@{}c@{}}c.r. $\geq 1$\\ \%\end{tabular}} \\ \toprule \midrule
\text{GPT-3, c.r. 60-80\%}                                                                                         & 82 (24)                                                             & \textbf{88}                                                                                    & 31                                \\
\text{GPT-3, c.r. 40-60\%}                                                                                         & 78 (28)                                                             & 81                                                                                                    & 33                                \\
\text{GPT-3, c.r. 20-40\%}                                                                                         & \textbf{59 (28)}                                                    & 76                                            & \textbf{11}                       \\ \midrule
Supervised baseline & 55 (29) & 71 & 7 \\ \midrule
\text{Referee-Distill, filter \#1}  & \text{\text{46 (15)}}          & \text{\text{89}}      & \text{\text{1}}             \\
\text{Referee-Distill, filter \#2} & \text{\text{49 (18)}}          & \text{91}     &  \text{\text{2}}             \\ \bottomrule

\end{tabular}
\caption{Statistics of automatically-generated datasets (ours and GPT-3). The first three rows refer to three different datasets generated through three-shot prompting GPT3-Instruct Curie with summaries from different compression ratios (c.r.).
Following rows show results for the third and last iteration of our models, using each one of the two described filters ($f_1$, $f_2$, see \S\ref{sec:filters}). Sentences correspond to a held-out set during training.
}\label{table:statistics-iterative}
\end{table} 

Besides using prompts that encourage shorter summaries, one can iteratively summarize through few-shot prompting. If $f_p(s)$ is the summary GPT-3 generates when prompted with $p$, then $f_p(f_p(\ldots f_p(s)))\!=\!f^n_p(s)$ may also be a summary of $s$, possibly shorter. 
We find that successive application of the same prompt did not result in shorter summaries, i.e. $|f^n_p(s)|\!\simeq\!|f_p(s)|$, suggesting $f_p$ is roughly idempotent in terms of length (see \ref{fig:idempotent}).

These experiments motivate the need of more sophisticated approaches to length control and reliably summarizing without supervision.

\section{Experiments}

\begin{table*}[t]
\footnotesize
  \centering
  \caption{Results for paired comparison between two models (``comparative'' and ``baseline'', see first two columns).
  We use BERT Score ($\in [0,1]$), and ROUGE-1,2,L ($\in [0, 100]$). Metric differences are computed as detailed in Section \ref{sec:gpt3-vs-distill}, and comparative model' score is shown in brackets. 
  A positive difference reflects an improvement using the comparative model over the baseline.
  The initial dataset consists of 10000 samples, from where we keep only summaries that differ in at most 10\% compression ratio, and are not identical.}\label{table-gpt3-vs-func1}
\begin{tabular}{@{}cccccccc@{}}
\toprule
\textit{\begin{tabular}[c]{@{}c@{}}Comparative \\ model\end{tabular}} & \textit{\begin{tabular}[c]{@{}c@{}}Baseline \\ model\end{tabular}} & \textit{\scriptsize\begin{tabular}[c]{@{}c@{}}BERTScore\\ F1 diff \\ {[}ours{]}\end{tabular}} & \textit{\scriptsize\begin{tabular}[c]{@{}c@{}}R-1 diff\\ {[}ours{]}\end{tabular}} & \textit{\scriptsize\begin{tabular}[c]{@{}c@{}}R-2 diff \\ {[}ours{]}\end{tabular}} & \textit{\scriptsize\begin{tabular}[c]{@{}c@{}}R-L diff \\ {[}ours{]}\end{tabular}} & \textit{\scriptsize \begin{tabular}[c]{@{}c@{}}\% \\ examples \\ to \\ compare \end{tabular}} & \textit{\scriptsize\begin{tabular}[c]{@{}c@{}}\% equal \\ summ.\end{tabular}} \\ \toprule \midrule
Iter. 3, filter \#1                                                                     & GPT3, 60-80\%                                                                                & \text{0.02 {[}0.77{]}}                                                            & \text{4.9 {[}59.1{]}}                                                & \text{8.8 {[}47.8{]}}                                                 & \text{6.6 {[}58.1{]}}                                                 & 10\%                                                                                        & 2\%                                                                    \\ 
Iter. 3, filter \#1                                                                     & GPT3, 40-60\%                                                                                  & \text{0.03 {[}0.76{]}}                                                            & \text{5.4 {[}56.9{]}}                                                & \text{5.4 {[}45.6{]}}                                                 & \text{7.2 {[}56.1{]}}                                                 & 14\%                                                                                        & 4\%                                                                    \\ 
Iter. 3, filter \#1                                                                     & GPT3, 20-40\%                                                                                & \text{0.07 {[}0.74{]}}                                                            & \text{12.7 {[}53.9{]}}                                               & \text{17.1 {[}42.4{]}}                                                & \text{15.2 {[}53.0{]}}                                                & 30\%                                                                                        & 4\%                                                                    \\ 
\midrule
Iter. 3, filter \#1                                                                     & Supervised baseline                                                                                & \text{0.10 {[}0.74{]}}                                                            & \text{14.0 {[}53.8{]}}                                               & \text{18.8 {[}42.4{]}}                                                & \text{15.4 {[}53.0{]}}                                                & 35\%                                                                                        & 0\%                                                                    \\ 
\midrule
Iter. 1, filter \#2     & Iter. 1, filter \#1                                                                                                           & 0.01 {[}0.83{]}                                                                  & 1.1 {[}69.5{]}                                                         & 2.5 {[}58.4{]}                                                          & 1.3 {[}68.0{]}                                                            & 35\%                                                                                      & 26\%                                                                   \\
Iter. 2, filter \#2     & Iter. 2, filter \#1                                                                                                              & 0.01 {[}0.80{]}                                                                   & 2.3 {[}63.3{]}                                                         & 4.7 {[}52.9{]}                                                          & 2.6 {[}62.3{]}                                                          & 36\%                                                                                      & 27\%                                                                   \\
Iter. 3, filter \#2     & Iter. 3, filter \#1                                                                                                   & \text{0.01 {[}0.75{]}}                                                         & \text{2.5 {[}55.4{]}}                                                & \text{4.9 {[}45.3{]}}                                                 & \text{2.7 {[}54.7{]}}                                                 & 38\%                                                                                      & 26\%                                                                   \\ \bottomrule
\end{tabular}
\end{table*} 





\paragraph{Dataset} 
We create the corpora $\mathcal{D}$ and $\mathcal{F}$ by sampling
contiguous sentence pairs from RealNews \citep{zellers2019defending} news articles.
We filter out sentences shorter than 50 characters. 
Using GPT-3 as the teacher, we summarize sentences in $\mathcal{D}_0$ and use the outputs with 60-80\% compression ratio as our initial dataset $\mathcal{C}_0$, since it was the best one quantitatively and qualitatively. 
Although this implies that we will not initially have enough short summaries to train from, we will generate progressively shorter ones during our training procedure. 

\paragraph{Training Details}
We use off-the-shelf model WANLI~\citep{liu2022wanli} to create the filter $f_\mathrm{NLI}$. We run 3 iterations of \firstphasemodelname{} with compression ratios for each training iteration are as follows: $r_1\!=\!0.7$, $r_2\!=\!0.5$, $r_3\!=\!0.3$. All generated data is decoded via beam search (with beam width 5). All experiments with $f_\mathrm{NSP}$ are done with $r=e^{-6} \approx 0.0025$, empirically decided through preliminary exploration. We train \secondphasemodelname{} with $n=10$ buckets for 7 iterations. For more details, refer to Appendix \ref{appendix:training_details}.

While having contextual information for $f_\mathrm{NSP}$ is useful, many applications do not have contextual information available. Therefore, we use the corpus $\mathcal{C}_{0} \cup \ldots \cup \mathcal{C}_{t}$ generated using $f_1$ as initial training data for \secondphasemodelname{} with the goal of increasing its applicability. 

\paragraph{Supervised Baseline} We include a supervised baseline as a comparison point. 
Following \citet{rush2015neural}, we use Gigaword \cite{gigaword, graff2003english} as a silver-labeled dataset where the headline is used as the summary for an article's first sentence. We fine-tune GPT2-Large \cite{radford2019language} (the same architecture as in our models) on this corpus, with \editms{default} hyperparameters. Due to the training data's nature, the supervised baseline has a low average compression ratio of 55\% (similar to \firstphasemodelname{} Iteration 2, but longer on average than Iteration 3). Importantly, we do not (and cannot) include a conventional knowledge distillation from GPT-3 (e.g. \citealt{shleifer2020pre}) since the full distribution of token logits is unavailable.

\subsection{Evaluating \firstphasemodelname{}}

\begin{figure}[]
    \centering
    \includegraphics[width=0.85\linewidth]{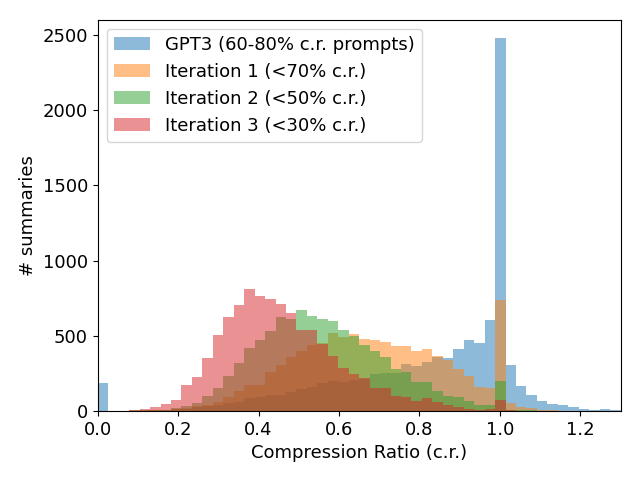}
    \caption{Histogram of compression ratios (c.r.) for $\mathcal{C}_0, \ldots, \mathcal{C}_3$ with filter \#1 ($f_1$, see \S\ref{sec:filters}). $\mathcal{C}_0$ are GPT-3 summaries, and $\mathcal{C}_1, \mathcal{C}_2, \mathcal{C}_3$ each of the training iterations.}
    \label{histogram-compression-ratio}
\end{figure}

\paragraph{Compression and Fidelity Statistics} We observe that iterative training with selection of progressively shorter summaries achieves the goal of generating shorter summaries, and additionally, less variance in compression ratios (Figure~\ref{histogram-compression-ratio}). Moreover, in Table \ref{table:statistics-iterative} we observe that, by using an NLI filter during distillation, \firstphasemodelname{} summaries were \textasciitilde90\% entailed by the original sentence according to WANLI {(compared to 79\% if not including an NLI filter during training)}. 
This vastly surpasses the comparable GPT-3 dataset (20-40\%), and achieves similar fidelity as the best GPT-3 summaries, even when our model is significantly smaller. The same trends hold for both filters.

\subsubsection{Comparison with GPT-3}\label{sec:gpt3-vs-distill}



We compare the quality of our summaries ($s'$) and GPT-3 generated summaries ($s''$), for every trained iteration and every GPT-3 dataset. Since longer summaries will naturally be able to preserve more of the original information \cite{schluter2017limits}, it is not reasonable to compare two wildly different compression ratios. Therefore, we only compare summaries that differ by length at most 10\%: $\big|\frac{|s'|}{|s|} - \frac{|s''|}{|s|}\big| \leq 0.1$, where $s$ is the original sentence.

To measure summary quality automatically, 
we compute BERTScore \cite{zhang2019bertscore} and ROUGE-1,2,L \cite{lin2004rouge} against the original sentence $s$ (no references are available). Given a metric $m$, we evaluate models based on $m(s', s) - m(s'', s)$. Positive values reflect that $s'$ had higher scores than the baseline summary $s''$, which is desirable for all our metrics.
Our models show significant improvements in all metrics when compared to every GPT-3 dataset, and the supervised baseline 
(see Table \ref{table-gpt3-vs-func1}). 
Our model shows especially large improvements when compared to the shortest GPT-3 summaries (20-40\% prompts). This suggests our iterative procedure was able to preserve quality better during the selection for shorter summaries.

Finally, we compare the effect of introducing a contextual filter $f_{\mathrm{NSP}}$. We compare two identical training runs that differ only in the filter applied (filter \#2 vs. \#1), and we observe small improvements when including $f_{\mathrm{NSP}}$ (see last rows of Table \ref{table-gpt3-vs-func1}, higher values mean \#2 was better than \#1).

\subsubsection{Human Evaluation}
\label{subsec:human_eval_distill}

\begin{table}[]
\centering
\footnotesize
\begin{tabular}{@{}cccc@{}}
\toprule
\textit{}                                                                      & {\footnotesize \textit{Faithful}} &  {\footnotesize \textit{Relevant}}  &  {\footnotesize \textit{Fluent}}  \\ \midrule
\text{\begin{tabular}[c]{@{}c@{}}Supervised Baseline\end{tabular}}  &  0.778 &  0.883  & 0.838 \\
\text{GPT-3, 20-40\%}   &  0.825 &  0.950 & 0.935 \\
\text{\begin{tabular}[c]{@{}c@{}}\textbf{Referee-Distill} \textbf{Iter. 3 (us)}\end{tabular}} & 0.835 & 0.963 & 0.915 \\
\bottomrule
\end{tabular}

\caption{Results for human evaluation on 100 samples in faithfulness, relevance, and fluency, all in $[0, 1]$ range (higher is better). See \S\ref{subsec:human_eval_distill} for all details.}
\vspace{-3mm}
\label{tab:human_distill}
\end{table}

\begin{comment}

\begin{table}[]
\centering
\footnotesize
\begin{tabular}{@{}cccc@{}}
\toprule
\textit{}                                                                      & Faithfulness & \begin{tabular}[c]{@{}c@{}}Information\\ Preservation\end{tabular} & Fluency \\ \toprule  \midrule
\textit{GPT-3 20-40\%}   &  0.825 &  0.950 & 0.935 \\ \midrule
\textit{\begin{tabular}[c]{@{}c@{}}Supervised\\ Baseline\end{tabular}}  &  0.778 &  0.883  & 0.838 \\
\midrule
\textit{\begin{tabular}[c]{@{}c@{}}Referee-Distill\\ Iteration 3\end{tabular}} & 0.835 & 0.963 & 0.915 \\
\bottomrule
\end{tabular}

\caption{\ms{TBD human evaluation for referee distill}}

\end{table}
\end{comment}

We conduct a human evaluation to verify the qualities of summaries from \firstphasemodelname{}. We measure 3 axes: faithfulness (\emph{is the summary true to the source?}), relevance (\emph{does the summary capture important information from the source?}) and fluency, each on a 3-point Likert scale.\footnote{Note that humans did not agree well between the top two buckets for relevance, so we combined these into a single bucket and calculate average and agreement on this.} We conduct our evaluation on 100 examples and find agreement by Fleiss $\kappa$ \citep{fleiss1971measuring} of 0.32, 0.34, and 0.57 (respectively) indicating fair to moderate agreement \citep{landis1977measurement}.

We compare between different methods to obtain succinct summaries: \firstphasemodelname{} (Iteration 3, function~\#1), GPT-3 20-40\%, and the supervised baseline. Following \S\ref{sec:gpt3-vs-distill}, we only compare sentences if all generations differ in compression ratio of at most 10\%. See results in Table~\ref{tab:human_distill}.

Broadly, we find that \firstphasemodelname{} and GPT-3 achieve significantly higher quality than the supervised baseline, with \firstphasemodelname{} showing even slightly higher scores than GPT-3 for all 3 axes.
We also note that this evaluation may somewhat favor baselines, since we only select examples where they achieved a similar compression ratio as us. We are not accounting for the fact that \firstphasemodelname{} can generate short summaries for many examples the other two systems cannot. 

\subsection{Evaluating \secondphasemodelname{}}


\begin{figure}[t]
\centering     
\subfigure[\secondphasemodelname{} bucket accuracy per iteration.]{\includegraphics[width=0.8\linewidth]{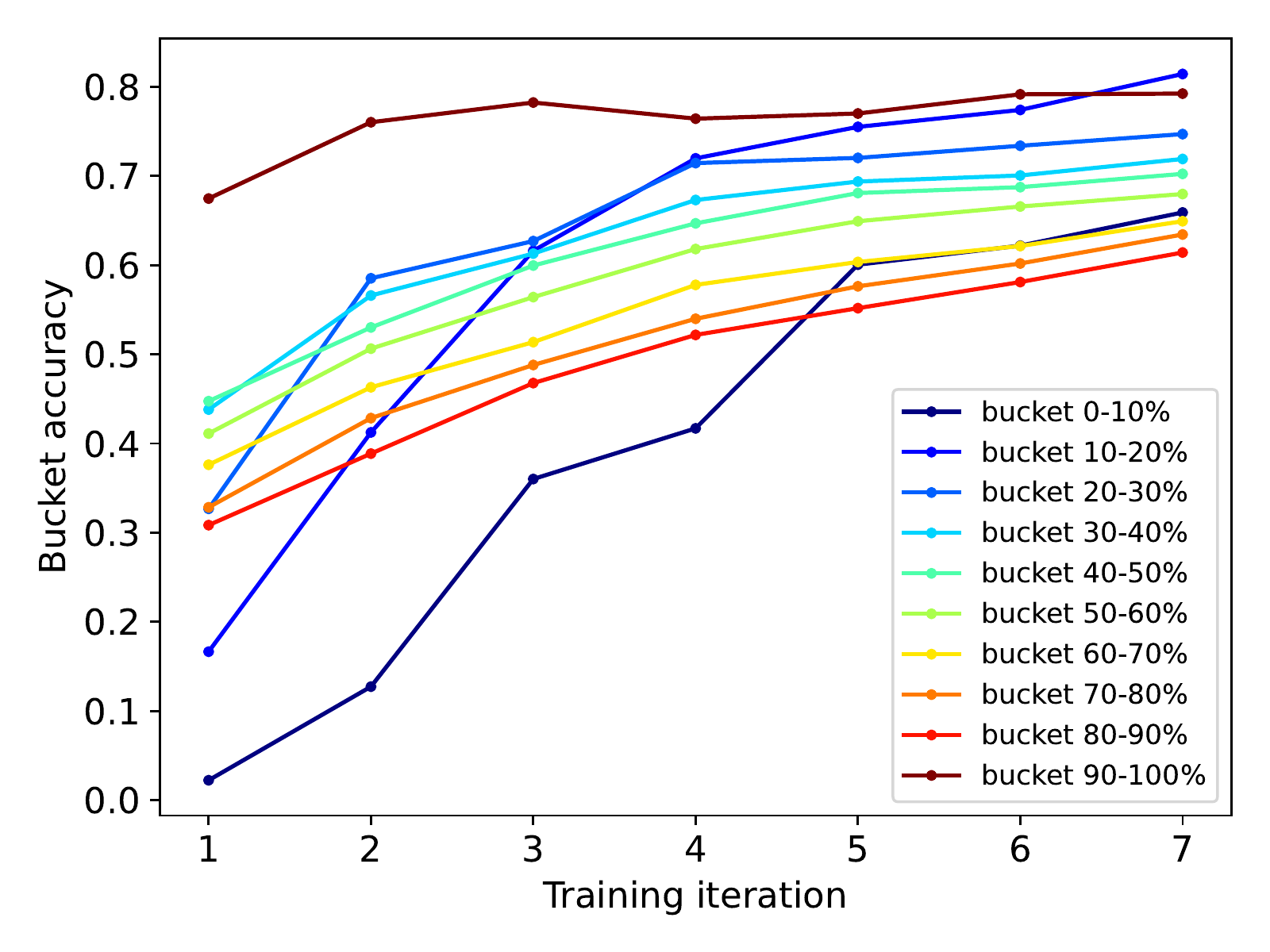}}
\subfigure[Compression ratio boxplot per iteration for bucket 30-40\%. Red lines represent bucket bounds, and blue lines represent a 10\% compression ratio difference to bucket bounds. Other buckets shown in Appendix \ref{appendix:control_code_boxplots}.]{\includegraphics[width=0.8\linewidth]{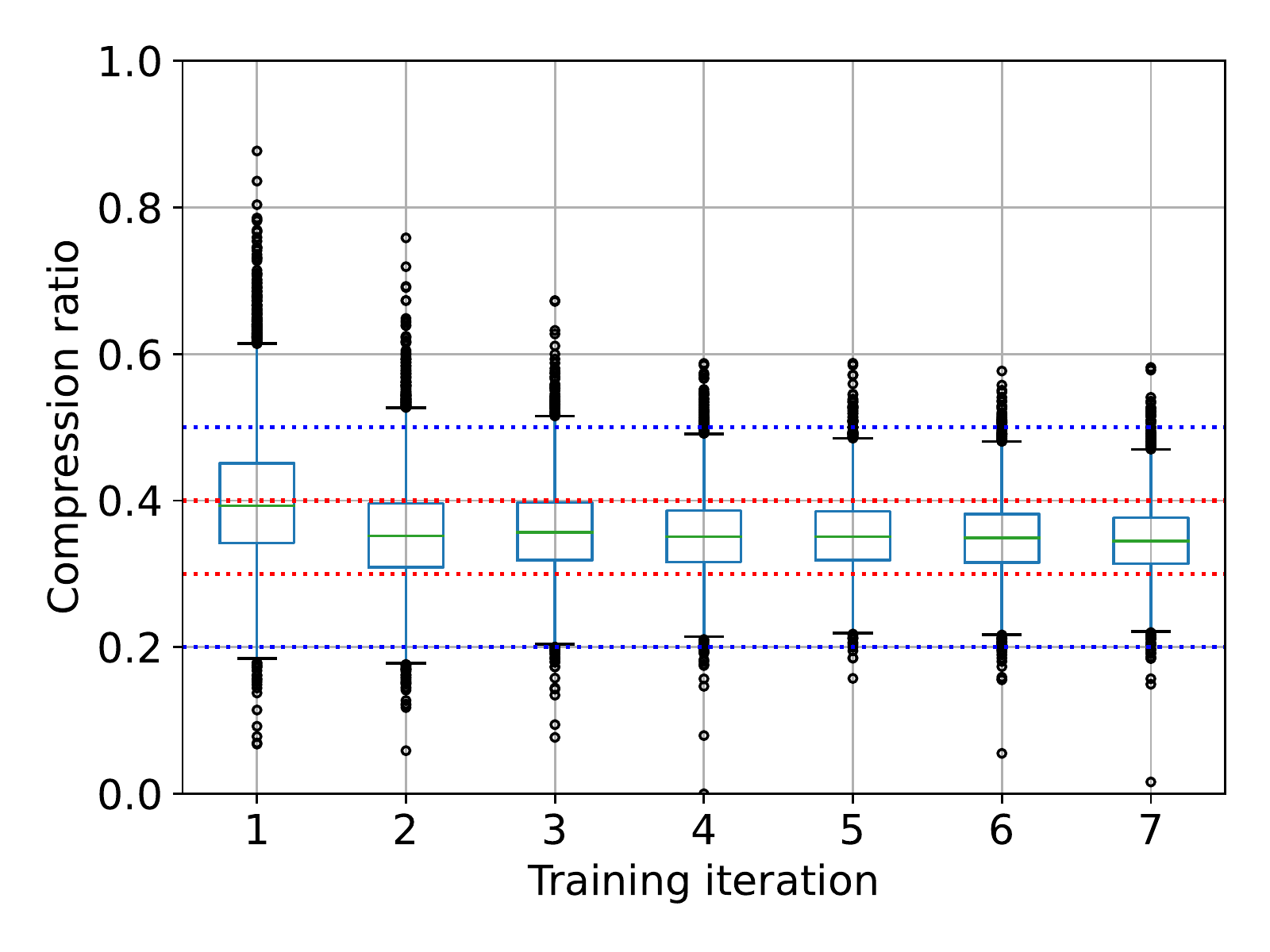}}
\caption{Statistics per \secondphasemodelname{} iteration, computed on the datasets $\mathcal{E}_i$ generated during training.}\label{fig:rainbow}
\vspace{-2mm}
\end{figure}


\begin{table}[]
\centering
\small
\begin{tabular}{@{}cccccc@{}}
\toprule
Model                               & \begin{tabular}[c]{@{}c@{}}BERT \\ Score \end{tabular} & R-1           & R-2           & R-L           & \begin{tabular}[c]{@{}c@{}}Bucket \\ Acc. \end{tabular} \\ \midrule
GPT3, 60-80\%          & 0.66                                                        & 40.0          & 24.3          & 36.8          & 3\%                                                       \\
GPT3, 40-60\%             & 0.65                                                        & 39.5          & 28.0          & 36.9          & 5\%                                                       \\
GPT3, 20-40\%          & 0.63                                                        & 34.1          & 18.4          & 31.0          & 14\%                                                     \\
\text{\scriptsize Referee-Ctrl}, Iter. 3 & \textbf{0.69}                                               & \text{47.7} & \textbf{36.7} & \text{47.3} & 61\%                                                      \\ 
\text{\scriptsize Referee-Ctrl}, Iter. 5 & \text{0.68}                                               & \text{46.5} & \text{34.8} & \textbf{46.0} & 69\%                                                      \\
\text{\scriptsize Referee-Ctrl}, Iter. 7 & 0.66                                                        & 44.5          & 32.1          & 43.9          & \textbf{71\%}                                             \\ \bottomrule

\end{tabular}
\caption{BERT Score, ROUGE-1,2,L and bucket accuracy for the 30-40\% bucket. Three GPT3 datasets are shown, and three different iterations of \secondphasemodelname{}. Data for other buckets is in Appendix \ref{appendix:control_code_tables}.}\label{table:bucket_stats} 
\end{table}


We train all our models with $n=10$ buckets to provide a very fine-grained control: the average sentence length in our dataset is 134 characters long, which implies that each bucket may span \textasciitilde13 characters for the average sentence. This implies a model may only have one or two words of freedom before exceeding the maximum length allowed in a bucket. It is important to note that models do not have direct access to a mapping of subword token to character length, and rather need to estimate character length during training. We use character length constrains since text applications will impose this type of restriction. 

We show that each \secondphasemodelname{} iteration increases bucket accuracy and reduces resulting compression rate variance (See Figure \ref{fig:rainbow}, Appendix \ref{appendix:control_code_boxplots}). To maximize quality, in all our experiments we only use one sampled beam. If we wished to maximize bucket accuracy at the expense of possibly reduced quality, we can take the top beams and select the most likely one that is in the prompted bucket. This procedure increases the bucket accuracy dramatically: in iteration two, using one beam has a \text{42\%} bucket accuracy for the bucket 80-90\% (in a held-out set), whereas using three yields an accuracy of \text{71\%}, and five, \text{82\%}. This same trend holds for other iterations and buckets, reaching \text{93\%} bucket accuracy in iteration 7.

This trade-off between bucket accuracy and summary quality that can be seen for bucket $b_3$ in Table \ref{table:bucket_stats}, although the behavior is consistent for all buckets (See Appendix \ref{appendix:control_code_tables}). There, Iteration 3 has slightly higher BERTScore and ROUGE than Iteration 7, at the expense of lower bucket accuracy. We believe this is because as bucket accuracy increases we reach harder to summarize examples at the desired length range, causing average scores to drop.

Later iterations also show more (small) disfluencies, which we partially attribute to the aforementioned cause. Also, small disfluencies may propagate over time, which we mitigate by using the mean negative log likelihood ratio filter $f_{\mathrm{AvgNLL}}$. Removing the fluency filter $f_{\mathrm{AvgNLL}}$ will also enable 9 more points of bucket accuracy on average.

\paragraph{Importance of Iterative Distillation} We trained a GPT2-Large model for 10 epochs with the same seed dataset $\mathcal{E}_0$, and compare with \secondphasemodelname{} iteration 5 (also trained for 10 epochs in total). Bucket accuracy of \secondphasemodelname{} was \textasciitilde20 points better: for 30-40\% bucket, bucket accuracy in the non-iterative version was 42\% vs. 69\% for \secondphasemodelname{}; for 70-80\% bucket, accuracy was 34\% vs. 58\%).




\subsubsection{Human Evaluation}
\label{subsec:human_eval_control}

We aim to explicitly test the capacity of systems to generate an \emph{acceptable summary.} That is, a summary to meets minimum human measures of quality, and also adheres to the desired length constraint. We omit the supervised baseline here, as it does not have explicit length control, and thus include only \secondphasemodelname{} and GPT-3.

Specifically, we measure summary accuracy as the fraction of summaries (of 100 randomly pre-selected sentences) that adhere to length control while being sufficiently fluent, relevant, and faithful. These axes are measured as in \S\ref{subsec:human_eval_distill}, achieving agreement by Fleiss $\kappa$ \citep{fleiss1971measuring} of 0.34, 0.22, and 0.25 (respectively) indicating fair agreement \citep{landis1977measurement}. We include two accuracy measurements: \emph{acc}, requiring adhering to length constraints as well as at least 2 (``fair'') out of 3 on all human measures of quality; and \emph{acc}$^+$, which requires 3 out of 3 on all measures along with length adherence.

\begin{table}[]
\centering
\footnotesize
\editms{
\begin{tabular}{lrr}
\toprule
Method &    \emph{acc} &   \emph{acc}$^+$ \\
\toprule \midrule
\multicolumn{3}{l}{\textbf{High compression (40-60\%)}} \\
\midrule
\secondphasemodelname{} Iter. 3 &  0.768 &  0.360 \\
GPT-3 40-60\%    &  0.194 &  0.095 \\
\citet{ghalandari2022efficient} & 0.530 &	0.153 \\
\citet{liu2022learning} & 0.320 &	0.043 \\
\citet{schumann2020discrete} & 0.357 & 	0.037\\
\midrule
\multicolumn{3}{l}{\textbf{Extreme compression (20-40\%)}} \\
\midrule
\secondphasemodelname{} Iter. 3 &  0.670 &  0.233 \\
GPT-3 20-40\%    &  0.250 &  0.117 \\
\citet{ghalandari2022efficient} & 0.431 & 0.081 \\
\citet{liu2022learning} & 0.333 & 0.034 \\
\citet{schumann2020discrete} & 0.337 & 0.048\\
\bottomrule
\end{tabular}
}
\caption{Accuracy of \editms{different models} at generating high-quality, length-controlled summaries {for two ranges, 20-40\% and 40-60\%}. Summaries must fulfill length constraints and meet human notions of quality at reasonable (\emph{acc}) or high (\emph{acc}\textsuperscript{+}) standards to be accepted. Higher is better. See setup details in \ref{sec:human_eval_referee_control}.}
\label{tab:human_control}
\end{table}

Table~\ref{tab:human_control} includes results for the 20-40\% and 40-60\% compression ranges, following our GPT-3 datasets (see \ref{sec:human_eval_referee_control} for more setup details). \editms{
\secondphasemodelname{} vastly outperforms GPT3 for both regimes and metrics. More precisely, for the 40-60\% regime \secondphasemodelname{} showed +296\% in \emph{acc} and +279\% in \emph{acc}\textsuperscript{+} when compared with GPT3; and for the 20-40\% regime, \secondphasemodelname{} obtained +68\% and +99\% respectively.} \editms{We additionally included three unsupervised summarization systems in the human evaluation. These systems perform some length control, making their summaries comparable, but all three models performed poorly when compared with \secondphasemodelname{}: they were at least 23 points below \secondphasemodelname{} in \emph{acc}, and 17 points below in \emph{acc}\textsuperscript{+}}.


Lastly, we would like to emphasize that \secondphasemodelname{} aims to summarize all examples at the requested compression, regardless of the original sentence's length or difficulty. GPT3, on the other hand, only summarizes in the requested compression for longer sentences, which generally correlate with easier cases (see details in \ref{sec:gpt3_distro_right_bucket}).

\section{Related Work}

\paragraph{Unsupervised Summarization} The vast majority of prior work in sentence summarization assumed access to large-scale text-summary paired datasets from which to train supervised models \citep{rush2015neural, nallapati-etal-2016-abstractive,narayan2018don}. Nonetheless, these datasets are costly to create, and naturally-occurring summarization datasets (such as news highlights) are noisy and not easily found in other domains. Therefore, recent work emphasized the need for developing unsupervised or self-supervised methods such as autoencoders \citep{miao2016language,baziotis-etal-2019-seq}, but in general they lead to less fluent summaries; more recent work has explored the Information Bottleneck Principle \citep{west-etal-2019-bottlesum} instead of the reconstruction loss of autoencoders.
Our work contributes to this emerging line of research by demonstrating an entirely different method based on symbolic knowledge distillation.

\paragraph{Length-Controlled Summarization}
While real world applications would require controlling for summary length, most prior work for automatic summarization has not proposed a principled mechanism for controlling the level of compression. Notable exceptions include \citet{kikuchi2016controlling} and \citet{he2020ctrlsum, fan2017controllable, liu2018controlling}. These last works developed supervised models for controllable length summarization by adding control codes that corresponded to a range of summary lengths---commonly referred to as \textit{buckets}. However, in both works the degree of control is heavily dependent on the training dataset, since bucket bounds are defined so that each one has the same number of examples; this may make one bucket correspond to a wide range of compression ratios. Our work adds a unique contribution by proposing a reference-free method that allows for full range of controls, and explicitly evaluates for that behavior.\\
\editms{Concurrently to this work, \citet{ghalandari2022efficient} and \citet{liu2022learning} (the latter optimizing \citealp{schumann2020discrete}) proposed unsupervised mechanisms that enforce length at the word level, either through a reward mechanism or with strict length enforcement. In contrast, our method uses iterative knowledge distillation to achieve length control. We also control length at the character level, rather than word level. This can result in a more fair and challenging notion of control, as it prevents performing the simple strategy in which all function words are removed first to maximize general meaning. Notably, and also in contrast to our work, both \citet{ghalandari2022efficient} and \citet{liu2022learning} require training separate models for compressing at different compression ratios.}

\paragraph{Knowledge Distillation} Many prior works have focused on similar notions of transferring knowledge between models through generation and distillation, and we draw particular inspiration from \citet{west2021symbolic}. 
\citet{shleifer2020pre} also follow a similar form to our work, distilling a summarizer from pretrained models. Our work differs in two key ways. 
First, like most distillation works, \citet{shleifer2020pre} assume having a model trained for the task and with access to its full distribution of token logits ---both not the case of GPT-3, used here. Second, like many distillation studies, \citet{shleifer2020pre} aim to mimic the teacher model's distribution, while we attempt to improve it.
This core detail sets us apart from many works employing a large teacher model \cite{kim-rush-2016-sequence,schick-schutze-2021-generating, Ye2022ZeroGenEZ}, teaching a student to mimic a distribution rather than improve it as in our case.

\paragraph{Natural Language Inference (NLI)  for Summarization} \citet{pasunuru2017towards, pasunuru2018multi, li2018ensure} have used NLI for summarization enhancement: \citet{pasunuru2017towards} use entailment in multi-task learning, and 
\citet{pasunuru2018multi, li2018ensure} use entailment probability as a reward. 
In this work, we propose an alternative approach for incorporating NLI for enhancing fidelity of summarization under the Symbolic Knowledge Distillation framework.

\begin{comment}
\begin{itemize}
    \item Unsupervised summarization
    \item Controlled-length summarization (e.g. by length or any other rule, like MuCoCo)
    \item Knowledge distillation
    \item READ THIS ONE: 
\end{itemize}
\subsection{Sentence Summarization}
\subsection{Controllable Summarization/Generation}
- https://arxiv.org/abs/1711.05217
\end{comment}

\section{Conclusions}
We presented \modelname{}, a framework for sentence summarization that can be trained without reference summaries, while allowing direct control for summary compression ratio. We uniquely proposed \emph{iterative} Symbolic Knowledge Distillation, where student models from the previous iteration of distillation serve as teacher models in the next. Distilled models are significantly smaller than the original teacher, GPT-3, and empirical results demonstrated that the final student models vastly outperform the much larger GPT3-Instruct model in terms of the controllability of compression ratios, without compromising the {resulting summaries' quality}. 
A useful by-product of this iterative distillation process is a high-quality sentence summarization dataset with varying degrees of compression, which we will release jointly with our models upon publication.

\editms{
\section*{Acknowledgements}
We thank Vidhisha Balachandran, Alisa Liu, and the whole Tsvetshop lab for the valuable discussions. We also thank OpenAI for providing academic access to the GPT-3 API.
This material is based upon work partly funded by the DARPA CMO under Contract No.~HR001120C0124, and by DARPA MCS program through NIWC Pacific (N66001-19-2-4031).
S.K. has been supported by a Google PhD Fellowship. 
Any opinions, findings and conclusions or recommendations expressed in this material are those of the authors and do not necessarily state or reflect those of the United States Government or any agency thereof.
}

\section*{Limitations}


\modelname{} was entirely developed and tested with sentences extracted from news articles. More work is needed to assess \modelname{}'s robustness when applying it to other domains. These domains may differ in text type, topic, or even temporal differences, that may cause a distribution shift.

\modelname{}'s success is also tied to other systems' quality, mainly the seed dataset generator (GPT-3) and the summary fidelity filter (operationalized using WANLI entailments). \modelname{} may propagate errors and biases in NLI entailments, which may be remedied in the future as NLI research progresses. We believe some edge cases generated by \modelname{} may be useful to further augment data in NLI systems, but that investigation was outside the scope of this paper.  

\modelname{} is built entirely at the sentence level, and more work is needed to extend it to paragraph or document-level, although some of the same ideas could be applied (e.g., control codes over longer inputs).


\bibliography{anthology}
\bibliographystyle{acl_natbib}

\clearpage
\appendix
\section{Further Insights on GPT-3 Datasets}
\subsection{Successive Application of GPT-3 Prompts}\label{fig:idempotent}
\begin{figure}[h]
    \centering
    \includegraphics[width=\linewidth]{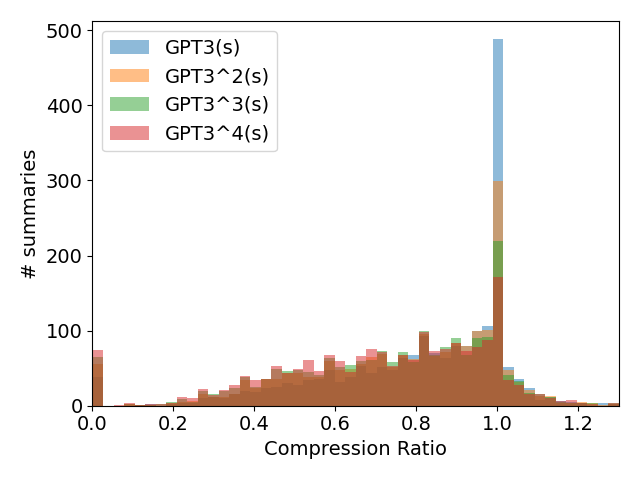}
    \caption{Histogram showing compression ratios of successive application of the same summarization prompts. Graphic shows no real difference in repeating the same prompt using the previously generated summary to summarize further.}
\end{figure}

\subsection{Compression Ratio Distribution per Dataset}
\begin{figure}[h]
    \centering
    \includegraphics[width=\linewidth]{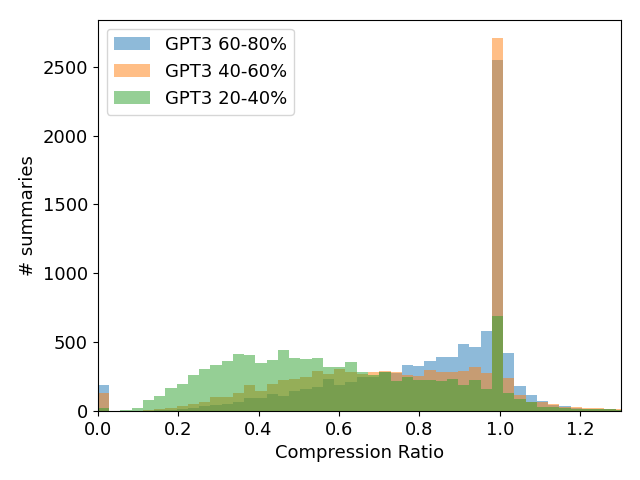}
    \caption{Histogram of compression ratio distribution for each prompt set.}
\end{figure}

\subsection{Distribution of Sentence Length in Cases Where GPT-3 Respected the Desired Compression}\label{sec:gpt3_distro_right_bucket}

\begin{figure}[h]
    \centering
    \includegraphics[width=\linewidth]{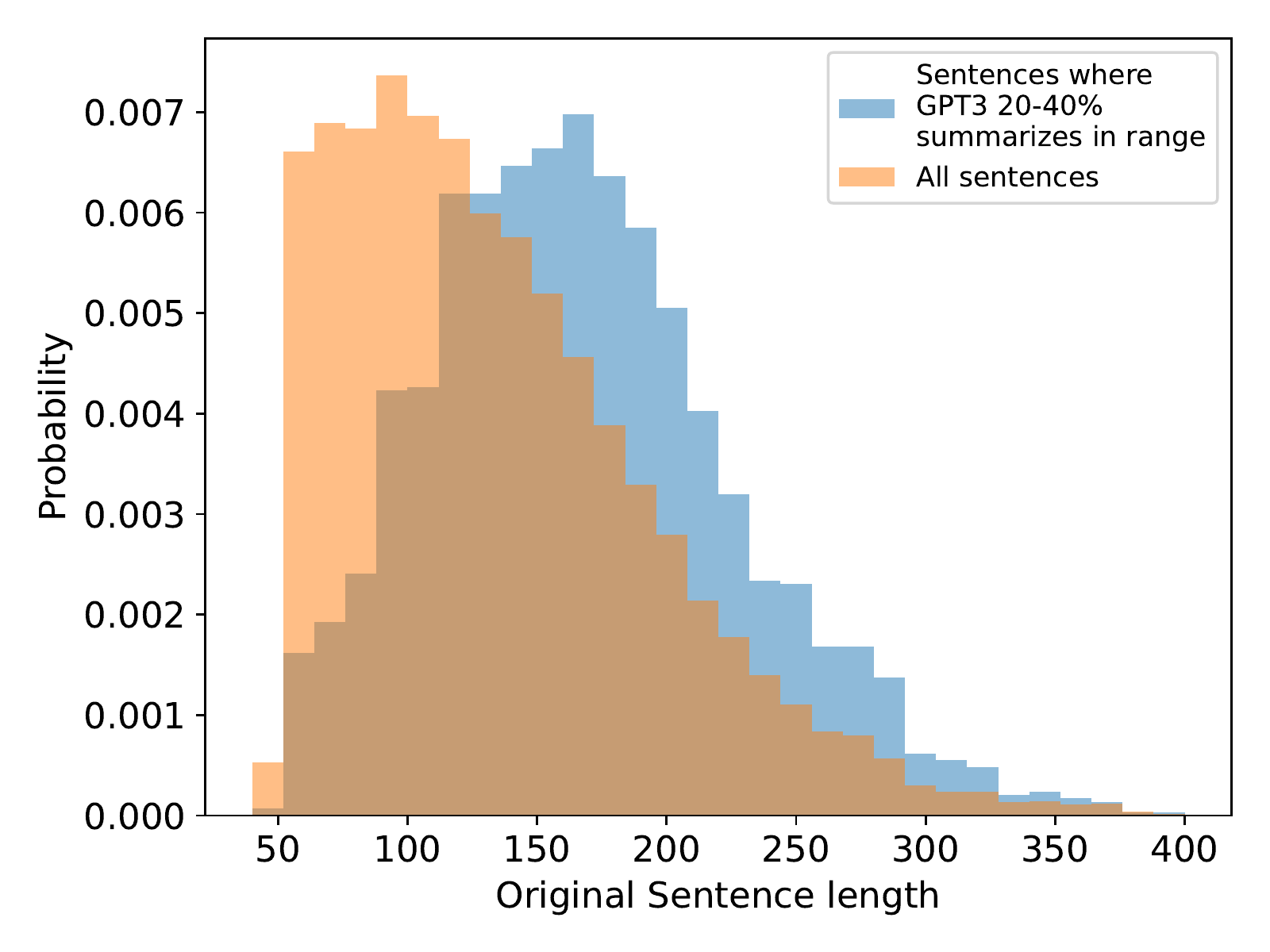}
    \caption{Distribution of original sentence length of all original sentences, and of the subset of samples where GPT-3 20-40\% returned a summary in the prompted compression range.}\label{fig:distro_orig_length_gpt3}
\end{figure}

\begin{figure}[h]
    \centering
    \includegraphics[width=\linewidth]{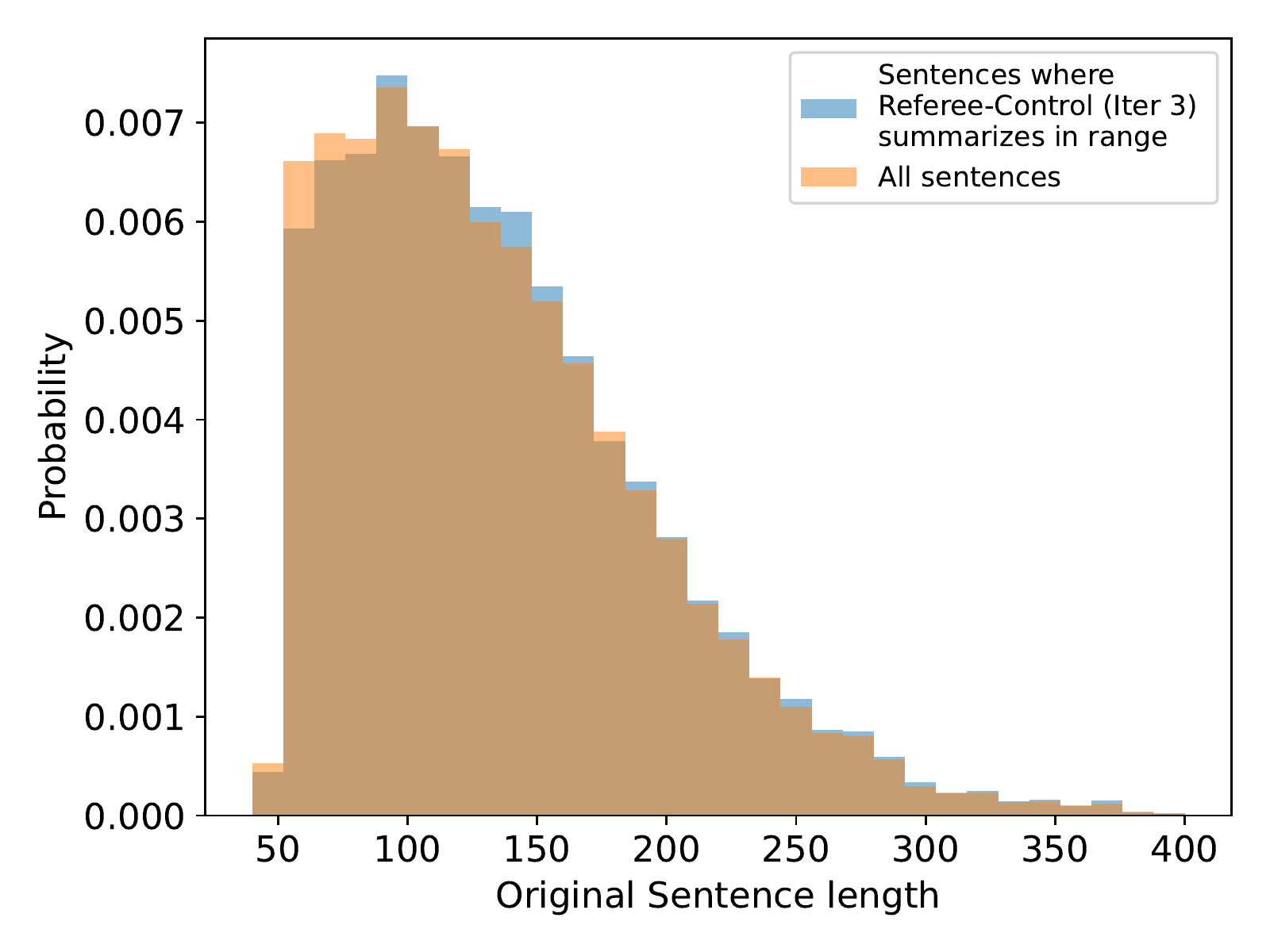}
    \caption{Distribution of original sentence length of all original sentences, and of the subset of samples where \secondphasemodelname{} 30-40\% returned a summary in the 20-40\% compression range (following the human evaluation setup).}\label{fig:distro_orig_length_referee}
\end{figure}

We observe that the distribution of original sentence length is markedly different when considering all the sentences in the dataset, versus when only considering the sentences where GPT-3's summary was in the desired compression range (see Figure \ref{fig:distro_orig_length_gpt3}. This shows that GPT-3's sentences compresses more only in easier cases, whereas \secondphasemodelname{} follows the distribution of the original set (see \ref{fig:distro_orig_length_referee}).

\section{Training and Dataset Details}\label{appendix:training_details}
We use GPT2-Large for all our fine-tuned models (774M weights, \textasciitilde
16x smaller than GPT3-Curie \citep{radford2019language,brown2020language}), and fine-tune for 5 epochs during each iteration of \firstphasemodelname{}, and for 2 epochs for \secondphasemodelname{}. $\mathcal{D}_0$ consists of 100000 sentences, $\mathcal{D}_{i>0}$ consists of 40000 sentences each; $\mathcal{F}_0$ consists of 220000 sentences, $\mathcal{F}_{i>0}$ consists of 10000 sentences each. All generated data is decoded with sampling beam search with 5 beams.

We balance each bucket before training each \secondphasemodelname{} iteration to avoid overrepresenting some classes. We use string characters 0 to $n-1$ as control codes for their respective buckets. We find that repeating the control code increases bucket accuracy --likely because the model attends more to these tokens--, so the end control code for bucket 2 will be \texttt{2 2 2 2 2 2 2 2 2 2} (ten repetitions).

\section{Additional Details for \secondphasemodelname{}}
\subsection{Performance for Additional Buckets}\label{appendix:control_code_tables}

\begin{table}[h]
\centering
\small
\begin{tabular}{@{}cccccc@{}}
\toprule
Model                               & \begin{tabular}[c]{@{}c@{}}BERT \\ Score \end{tabular} & R-1           & R-2           & R-L           & \begin{tabular}[c]{@{}c@{}}Bucket \\ Acc. \end{tabular} \\ \midrule
GPT3, 60-80\%          & 0.76                                                        & 57.5          & 41.9          & 54.6          & 7\%                                                       \\
GPT3, 40-60\%             & 0.76                                                        & 58.9          & 48.1          & 56.6          & 10\%                                                       \\
GPT3, 20-40\%          & 0.71                                                        & 48.9          & 32.1          & 44.8          & 13\%                                                     \\
\text{\scriptsize Referee-Ctrl}, Iter. 3 & \text{0.79}                                               & \text{65.1} & \text{54.9} & \text{64.5} & 56\%                                                      \\ 
\text{\scriptsize Referee-Ctrl}, Iter. 5 & \text{0.79}                                               & \text{64.7} & \text{54.1} & \text{64.2} & 66\%                                                      \\
\text{\scriptsize Referee-Ctrl}, Iter. 7 & 0.78                                                        & 63.2          & 51.9          & 62.7          & \text{68\%}                                             \\ \bottomrule

\end{tabular}
\caption{BERT Score, ROUGE-1,2,L and bucket accuracy for the 50-60\% bucket. Three GPT3 datasets are shown, and three different iterations of \secondphasemodelname{}.}
\end{table}

\begin{table}[h]
\centering
\small
\begin{tabular}{@{}cccccc@{}}
\toprule
Model                               & \begin{tabular}[c]{@{}c@{}}BERT \\ Score \end{tabular} & R-1           & R-2           & R-L           & \begin{tabular}[c]{@{}c@{}}Bucket \\ Acc. \end{tabular} \\ \midrule
GPT3, 60-80\%  & 0.84                 & 72.3               & 58.1               & 70.0               & 11\%                 \\
GPT3, 40-60\%  & 0.83                 & 71.6               & 62.4               & 69.3               & 10\%                 \\
GPT3, 20-40\%  & 0.79                 & 62.4               & 46.9               & 58.3               & 8\%                  \\
\text{\scriptsize Referee-Ctrl}, Iter. 3 & 0.88                 & 79.3               & 71.4               & 78.7               & 48\%               \\
\text{\scriptsize Referee-Ctrl}, Iter. 5 & 0.88                 & 78.9               & 70.9               & 78.5               & 57\%                 \\
\text{\scriptsize Referee-Ctrl}, Iter. 7 & 0.87                 & 77.6               & 68.9               & 77.3               & 63\% \\ \bottomrule
\end{tabular}
\caption{BERT Score, ROUGE-1,2,L and bucket accuracy for the 70-80\% bucket. Three GPT3 datasets are shown, and three different iterations of \secondphasemodelname{}.}
\end{table}

It is crucial to note that because variance reduces with the number of iterations, cases where the summary did not land on the requested compression ratio are close to fulfilling the constraint. For example, when prompting with 30-40\%, 87.5\% of the samples fell in the 20-40\% range for \secondphasemodelname{} Iteration 7 (77\% for Iteration 3; 83\% for Iteration 5).

\newpage
\subsection{\secondphasemodelname{} Compression Ratio Distribution per Training Iteration}\label{appendix:control_code_boxplots}
\begin{figure}[!h]
\centering
\subfigure[Bucket 10-20\%]{\includegraphics[width=0.73\linewidth]{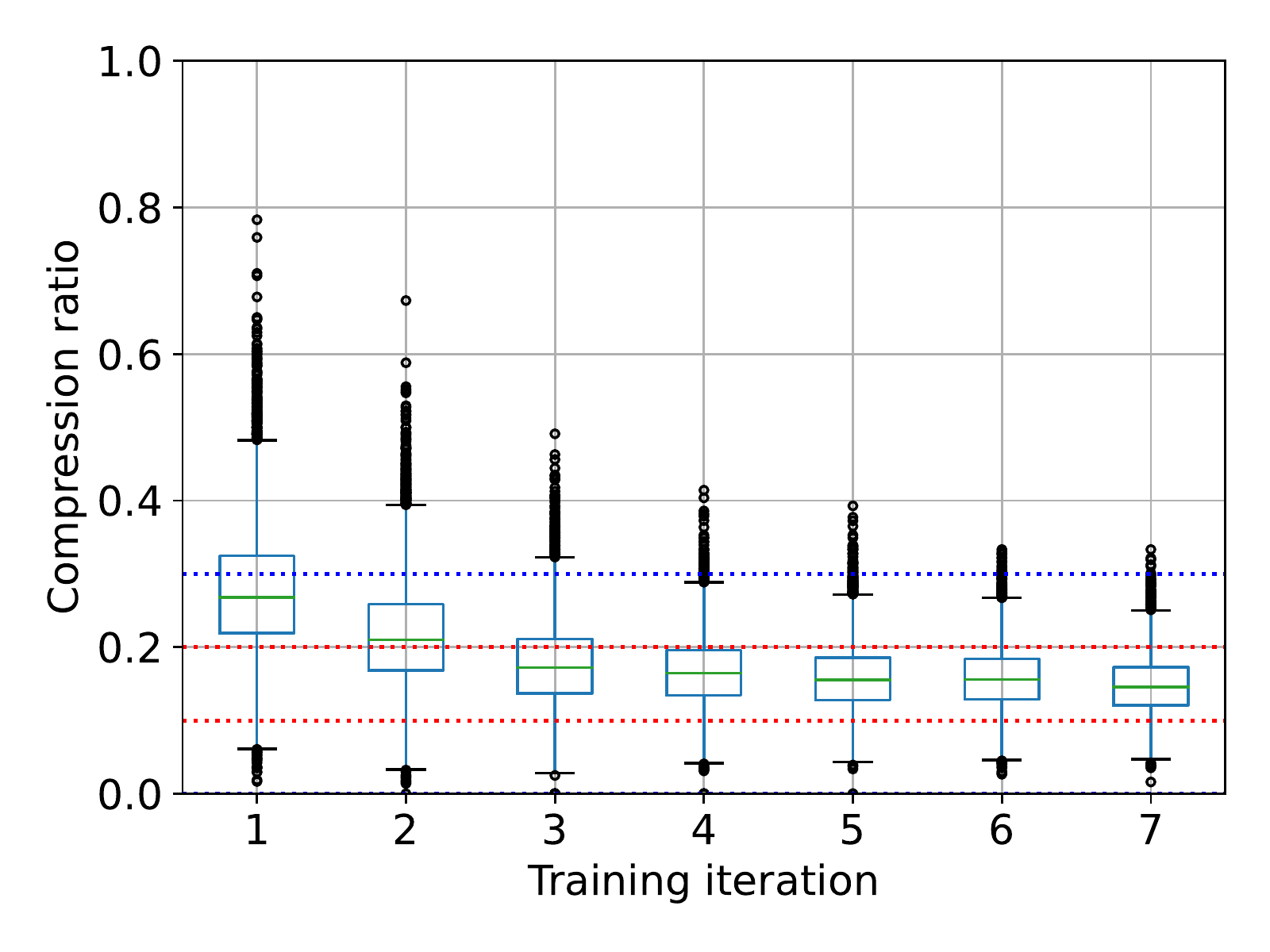}}
\subfigure[Bucket 40-50\%]{\includegraphics[width=0.73\linewidth]{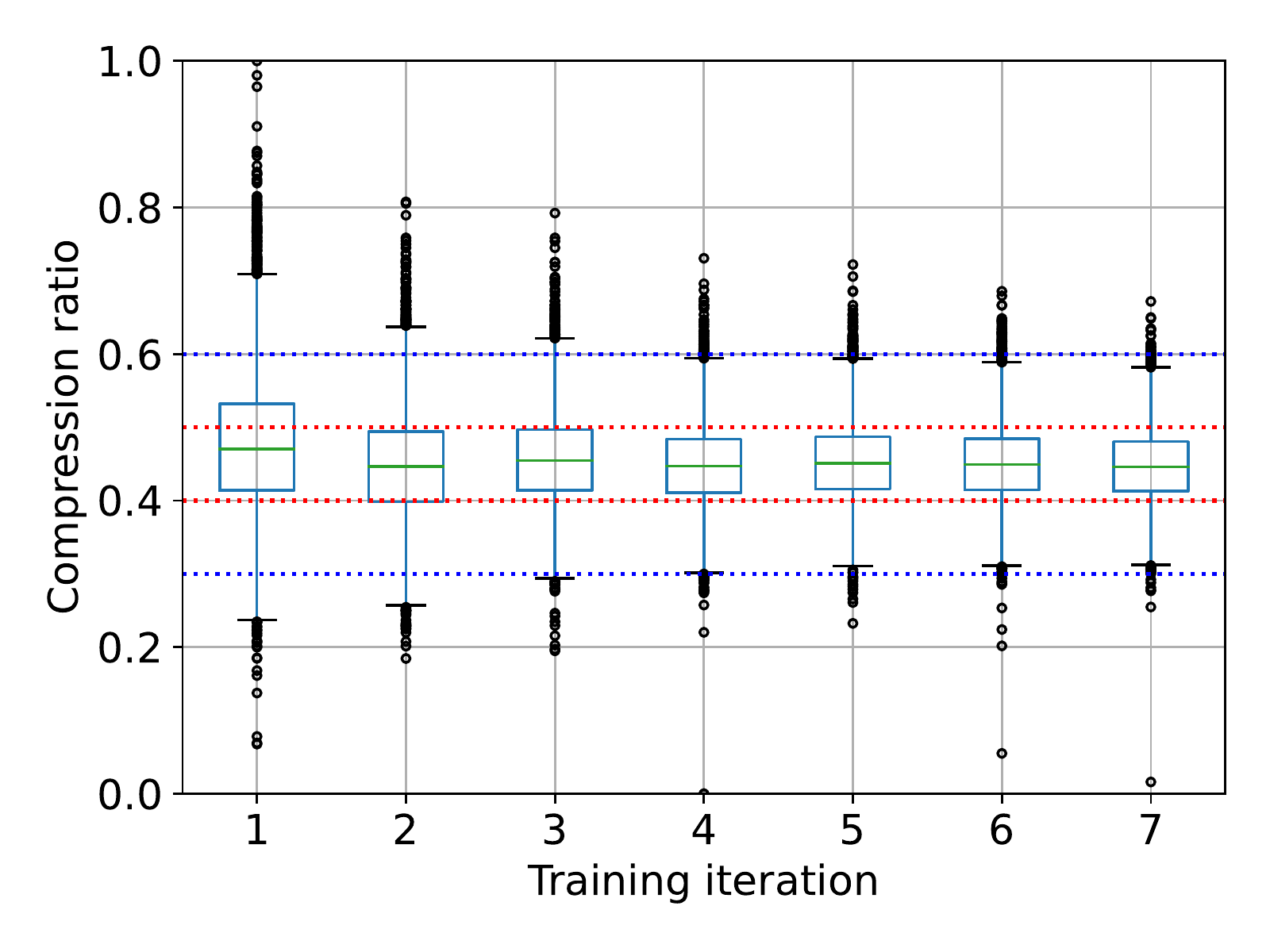}}
\subfigure[Bucket 50-60\%]{\includegraphics[width=0.73\linewidth]{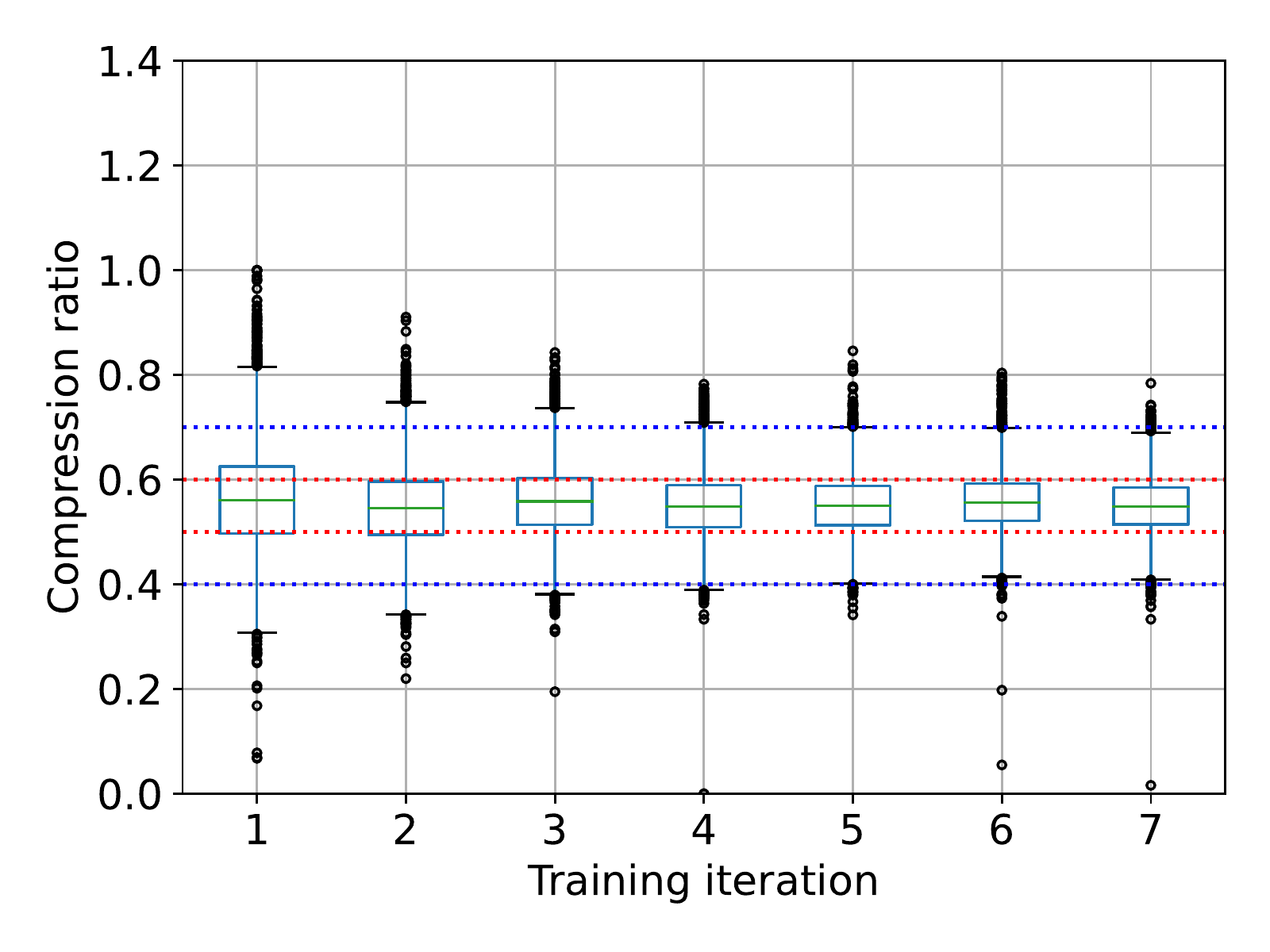}}
\subfigure[Bucket 70-80\%]{\includegraphics[width=0.73\linewidth]{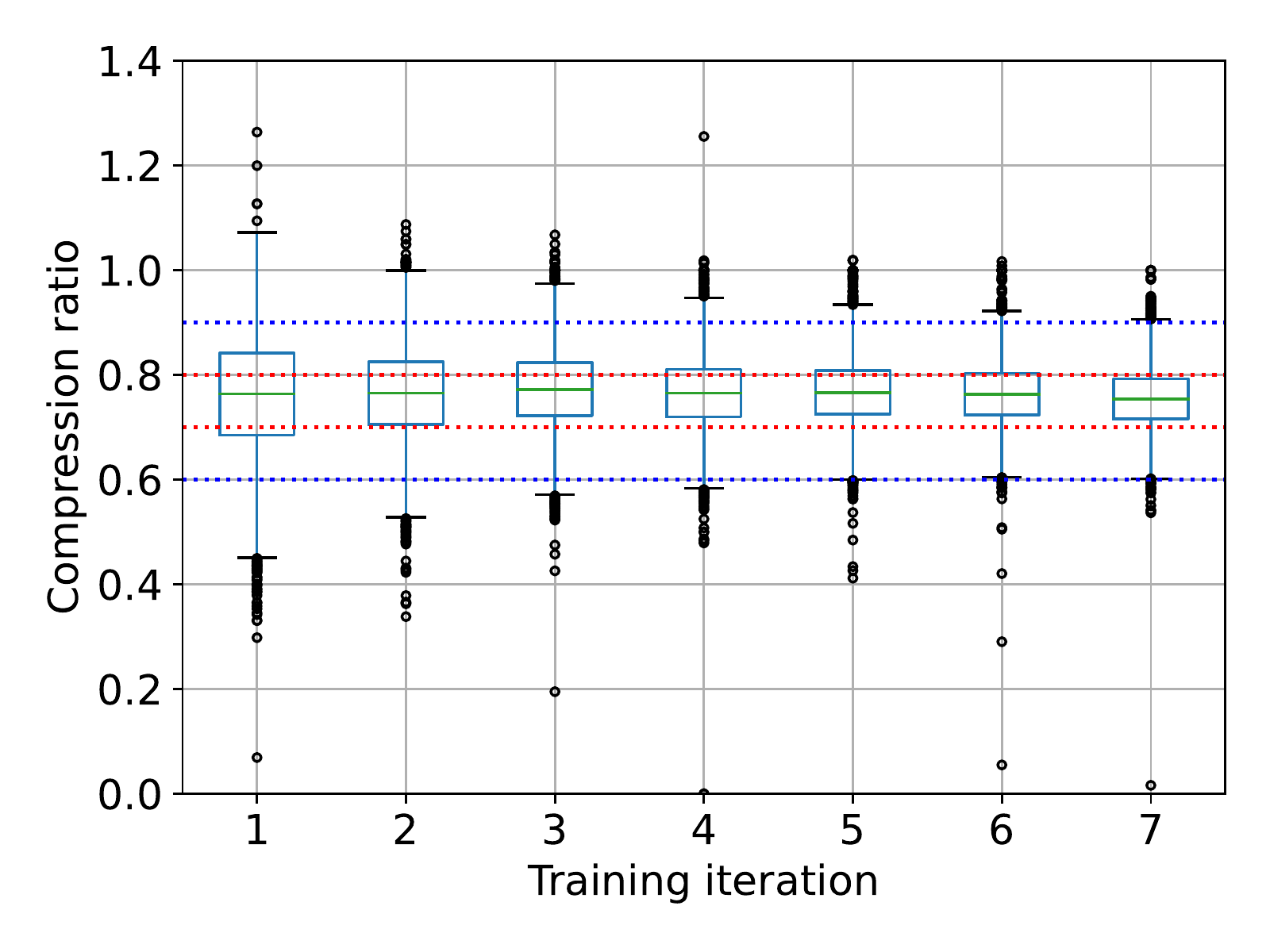}}
\caption{Boxplots showing the compression ratio distribution per training iteration, per bucket. 30-40\% bucket is shown in Figure \ref{fig:rainbow} in the main paper.}
\end{figure}

\subsection{Human Evaluation Setup Details}\label{sec:human_eval_referee_control}

For the 20-40\% range, we compare GPT-3 against \secondphasemodelname{} conditioned for 30-40\%, and accept all summaries within the same range as GPT-3 prompts (20-40\%). Similarly, for 40-60\% we condition \secondphasemodelname{} with 50-60\%, and accept all summaries within 40-60\% to equalize comparisons.

\editms{Regarding baselines, we trained \citet{ghalandari2022efficient} from scratch using RealNews. \citet{schumann2020discrete} was used out-of-the-box, and we use the default training data for \citet{liu2022learning} --the publicly available \citet{schumann2020discrete} generations-- due to time constraints. All baselines measure compression ratio at the word level, in contrast to our work that does it at the character level.
We train with a 30\% target compression ratio in baselines when comparing with the 20-40\% regime, and 50\% when comparing with the 40-60\% regime. This is done to maximize the number of examples that fall in the desired range, since some baselines enforce target compression ratio more strictly than others.
}

\section{On BERT Score Monotonicity}
In \S\ref{sec:dual}, we discussed that introducing control codes allows us to solve a related question: given we have a specific tolerance for information loss $k$, what is the shortest summary we could write? Estimating the level of information preservation with BERT Score Recall, we could sample from each bucket and select the shortest summary with BERT Score Recall $\geq k$. BERT Score Recall scores are generally well-ordered when ordering increasingly by bucket: we see that on average, the longest non-decreasing subsequence of BERT Score Recall scores is \textasciitilde7.7-7.8 for all iteration steps. Having a non-decreasing subsequence of 10 would mean that BERT Scores are perfectly ordered (a summary in a higher bucket would have always higher BERT Score recall than a lower bucket). 7.8 means that scores are generally well-ordered, with some noisy generations. 

\editms{\section{Examples of Generations}}

\subsection{Random sample of generations used in the 40-60\% human evaluation}

\noindent\textit{Original:} Viggo left South America aged 11, when his parents divorced and he moved to upstate New York with his mother and siblings. \\
\textit{GPT3:} Viggo left South America when his parents divorced. \textbf{[41.8\%, in range]}\\
\textit{\secondphasemodelname{}:} Viggo left South America aged 11, when his parents divorced him. \textbf{[52.46\%, in range]}\\
\textit{\citet{ghalandari2022efficient}:} Viggo left South America aged 11 when his parents divorced. \textbf{[48.36\%, in range]}\\
\textit{\citet{liu2022learning}:} parents divorced moved to upstate new york with his mother and siblings \textbf{[58.2\%, in range]}\\

\noindent\textit{Original:} And Yahoo appears to have had another money-losing fumble on its hands with the free livestream of the NFL’s Oct. 25 Bills-Jaguars game from London. \\
\textit{GPT3:} Yahoo appears to have had another money-losing fumble on its hands with the free livestream of the NFL’s Oct. 25 Bills-Jaguars game from London. \textbf{[97.3\%, 37.3\% too long]}\\
\textit{\secondphasemodelname{}:} Yahoo appears to have had another money-losing fumble on its hands with the free livestream. \textbf{[62.16\%, 2.16\% too long]}\\
\textit{\citet{ghalandari2022efficient}:} And Yahoo appears to have had another money-losing fumble with livestream Oct. 25 Bills-Jaguars. \textbf{[64.86\%, 4.86\% too long]}\\
\textit{\citet{liu2022learning}:} and yahoo have hands of nfl game \textbf{[21.62\%, 18.38\% too short]}\\

\noindent\textit{Original:} Suárez almost levelled with his latest sweeping free-kick from distance but the clearer chances came to Hull against a ponderous Liverpool defence. \\
\textit{GPT3:} Suarez almost levelled with his latest sweeping free-kick from distance but Hull's chances came against a ponderous Liverpool defense. \textbf{[91.16\%, 31.16\% too long]}\\
\textit{\secondphasemodelname{}:} Suarez almost levelled with his latest sweeping free-kick from distance. \textbf{[48.98\%, in range]}\\
\textit{\citet{ghalandari2022efficient}:} Suárez almost levelled with his free-kick to Hull Liverpool. \textbf{[40.82\%, in range]}\\
\textit{\citet{liu2022learning}:} <|unk|> from distance the clearer chances came to hull liverpool <|unk|> \textbf{[48.98\%, in range]}\\

\noindent\textit{Original:} Specifically, the company agreed to limit pre-payments, to provide accurate estimates of charges, and to disclose details of financing agreements. \\
\textit{GPT3:} Verizon has agreed to limit pre-payments, to provide accurate estimates of charges, and to disclose details of financing agreements. \textbf{[90.41\%, 30.41\% too long]}\\
\textit{\secondphasemodelname{}:} The company agreed to limit pre-payments and to provide accurate estimates of charges. \textbf{[58.9\%, in range]}\\
\textit{\citet{ghalandari2022efficient}:} Specifically, the company agreed to limit pre-payments to estimates financing. \textbf{[53.42\%, in range]}\\
\textit{\citet{liu2022learning}:} company agreed to limit to provide accurate estimates of disclose details financing \textbf{[56.85\%, in range]}\\

\noindent\textit{Original:} "There was a family in the car, they got out before it hit. \\
\textit{GPT3:} There was a family in the car when the car hit a tree, and they all got out before it hit the tree. \textbf{[167.8\%, 107.8\% too long]}\\
\textit{\secondphasemodelname{}:} There was a family in the car. \textbf{[50.85\%, in range]}\\
\textit{\citet{ghalandari2022efficient}:} "There was a family in the car. \textbf{[52.54\%, in range]}\\
\textit{\citet{liu2022learning}:} there was a family they got before it \textbf{[62.71\%, 2.71\% too long]}\\

\noindent\textit{Original:} As for the coming revolution, I can’t say for certain when that will happen. \\
\textit{GPT3:} I can't say for certain when the coming revolution will happen. \textbf{[82.89\%, 22.89\% too long]}\\
\textit{\secondphasemodelname{}:} I can't say for certain when it will happen. \textbf{[57.89\%, in range]}\\
\textit{\citet{ghalandari2022efficient}:} As for the coming revolution I can't say. \textbf{[53.95\%, in range]}\\
\textit{\citet{liu2022learning}:} for the coming revolution i can <|unk|> t say \textbf{[59.21\%, in range]}\\

\noindent\textit{Original:} In one New Jersey county, the Red Cross was AWOL, says Lt. Matthew Tiedemann of the Bergen County Office of Emergency Management. \\
\textit{GPT3:} In one New Jersey county, the Red Cross was AWOL. \textbf{[37.98\%, 2.02\% too short]}\\
\textit{\secondphasemodelname{}:} In one New Jersey county, the Red Cross was AWOL, says Lt. Matthew Tiedemann. \textbf{[59.69\%, in range]}\\
\textit{\citet{ghalandari2022efficient}:} In one New Jersey county Red Cross was AWOL says Lt. Matthew Tiedemann. \textbf{[55.04\%, in range]}\\
\textit{\citet{liu2022learning}:} in new jersey county red cross awol lt. <|unk|> emergency management \textbf{[52.71\%, in range]}\\

\noindent\textit{Original:} Furthermore, though some might argue that the purpose of food is merely to remind us of our own mortality, I would argue that the best cuisine can be an art form. \\
\textit{GPT3:} Food can be both utilitarian and artful. \textbf{[24.69\%, 15.31\% too short]}\\
\textit{\secondphasemodelname{}:} Furthermore, I would argue that the best cuisine can be an art form, rather than a food source. \textbf{[58.64\%, in range]}\\
\textit{\citet{ghalandari2022efficient}:} Furthermore of food mortality I would argue that the best cuisine can be an art form. \textbf{[52.47\%, in range]}\\
\textit{\citet{liu2022learning}:} the is our would that the best cuisine can be an art \textbf{[32.1\%, 7.9\% too short]}\\

\noindent\textit{Original:} Without proper conservation facilities nearby, the artifacts Batchvarov and his colleagues discovered had to be reburied underwater. \\
\textit{GPT3:} Without proper conservation facilities nearby, the artifacts Batchvarov and his colleagues discovered had to be reburied underwater. \textbf{[100.0\%, 40.0\% too long]}\\
\textit{\secondphasemodelname{}:} The artifacts Batchvarov and his colleagues discovered had to be reburied. \textbf{[56.06\%, in range]}\\
\textit{\citet{ghalandari2022efficient}:} Without conservation artifacts Batchvarov had to reburied underwater \textbf{[51.52\%, in range]}\\
\textit{\citet{liu2022learning}:} conservation facilities artifacts colleagues discovered to be reburied underwater \textbf{[61.36\%, 1.36\% too long]}\\

\noindent\textit{Original:} Often people are desperate or ridiculous rather than heroic, so I do what needs to be done.” \\
\textit{GPT3:} That person is heroic. \textbf{[23.91\%, 16.09\% too short]}\\
\textit{\secondphasemodelname{}:} Often people are desperate or ridiculous, so I do it. \textbf{[57.61\%, in range]}\\
\textit{\citet{ghalandari2022efficient}:} Often people are desperate or ridiculous heroic I. \textbf{[54.35\%, in range]}\\
\textit{\citet{liu2022learning}:} so i do what needs to be done <|unk|> \textbf{[40.22\%, in range]}\\

\noindent\textit{Original:} Liam’s feet make a guest appearance in the picture, as does his Great Dane Watson who he bought with ex-girlfriend Sophia Smith can be seen by his settee. \\
\textit{GPT3:} Liam's feet and his Great Dane Watson can be seen by his settee in the picture. \textbf{[51.3\%, in range]}\\
\textit{\secondphasemodelname{}:} Liam's feet make a guest appearance in the picture, as does his Great Dane Watson. \textbf{[53.25\%, in range]}\\
\textit{\citet{ghalandari2022efficient}:} Liam’s feet make a guest appearance in the picture as Great Dane Watson ex-girlfriend settee. \textbf{[60.39\%, 0.39\% too long]}\\
\textit{\citet{liu2022learning}:} <|unk|> feet picture does dane watson he bought with ex-girlfriend sophia smith be <|unk|> \textbf{[58.44\%, in range]}\\

\noindent\textit{Original:} I like this little gallery, the Small Weston Room, where the paintings go right up to the ceiling. \\
\textit{GPT3:} I like this little gallery, the Small Weston Room. \textbf{[51.02\%, in range]}\\
\textit{\secondphasemodelname{}:} I like this little gallery, the Small Weston Room. \textbf{[51.02\%, in range]}\\
\textit{\citet{ghalandari2022efficient}:} I like this little gallery Small Weston Room where paintings ceiling. \textbf{[70.41\%, 10.41\% too long]}\\
\textit{\citet{liu2022learning}:} i like room where the go right to the ceiling \textbf{[45.92\%, in range]}\\

\noindent\textit{Original:} After a stint as a lobbyist, during which he helped establish two think tanks affiliated with James Dobson’s hugely influential Focus on the Family, Parker became a confidant of Roy Moore, then a county judge. \\
\textit{GPT3:} After a stint as a lobbyist, during which he helped establish two think tanks affiliated with James Dobson's hugely influential Focus on the Family, Parker became a confidant of Roy Moore, then a county judge. \textbf{[100.0\%, 40.0\% too long]}\\
\textit{\secondphasemodelname{}:} After a stint as a lobbyist, Parker became a confidant of Roy Moore, then a county judge, during a stint. \textbf{[50.24\%, in range]}\\
\textit{\citet{ghalandari2022efficient}:} After a stint as a lobbyist helped think tanks James Dobson, Parker became a confidant of Roy Moore. \textbf{[47.85\%, in range]}\\
\textit{\citet{liu2022learning}:} a a lobbyist s focus on the family parker became confidant roy moore \textbf{[32.54\%, 7.46\% too short]}\\

\noindent\textit{Original:} Germany will be the favourites but it is wide open between the other three teams - Poland and Ukraine are a similar standard to Romania and Hungary, who we came up against in qualifying. \\
\textit{GPT3:} Germany will be the favourites, but it is wide open between the other three teams. \textbf{[44.09\%, in range]}\\
\textit{\secondphasemodelname{}:} Germany will be the favourites, but it is wide open between the other three teams. \textbf{[44.09\%, in range]}\\
\textit{\citet{ghalandari2022efficient}:} Germany will be the favourites but it is wide open between the other three teams - Poland and Ukraine. \textbf{[54.84\%, in range]}\\
\textit{\citet{liu2022learning}:} wide between the other three teams poland and are romania hungary we came \textbf{[39.25\%, 0.75\% too short]}\\

\noindent\textit{Original:} It is the latest mass killing attributed to the extremists, activists said, with the violence coming during a 24-hour period. \\
\textit{GPT3:} The latest mass killing attributed to the extremists occurred during a 24-hour period. \textbf{[68.8\%, 8.8\% too long]}\\
\textit{\secondphasemodelname{}:} It is the latest mass killing, activists said, with the violence coming. \textbf{[57.6\%, in range]}\\
\textit{\citet{ghalandari2022efficient}:} It is the latest mass killing attributed to the extremists 24-hour. \textbf{[53.6\%, in range]}\\
\textit{\citet{liu2022learning}:} is the latest mass killing attributed to the extremists \textbf{[44.0\%, in range]}\\

\noindent\textit{Original:} The brutal honesty and reality of death speaks to the truth and tragedy of bad timing for these two characters.” \\
\textit{GPT3:} The brutal honesty and reality of death speaks to the truth and tragedy of bad timing for these two characters. \textbf{[99.11\%, 39.11\% too long]}\\
\textit{\secondphasemodelname{}:} The brutal honesty and reality of death speaks to the truth. \textbf{[53.57\%, in range]}\\
\textit{\citet{ghalandari2022efficient}:} The brutal honesty and of death speaks to timing characters. \textbf{[53.57\%, in range]}\\
\textit{\citet{liu2022learning}:} the reality of death the truth and of bad timing characters \textbf{[52.68\%, in range]}\\

\noindent\textit{Original:} It's a nice touch of confectionerial twinning, and broadly chimes with the optional theme of this year's show, 'Man Made', which judging by the parrots and skyscapes and severed heads and moths has not been taken up with widespread enthusiasm. \\
\textit{GPT3:} The show's optional theme is 'Man Made'. \textbf{[16.46\%, 23.54\% too short]}\\
\textit{\secondphasemodelname{}:} It's a nice touch of confectionery and broadly chimes with the optional theme of this year's show, "Man Made", which has not been taken up. \textbf{[57.2\%, in range]}\\
\textit{\citet{ghalandari2022efficient}:} It's a nice touch of confectionerial twinning and chimes with theme of this year's show 'Man Made parrots. \textbf{[43.62\%, in range]}\\
\textit{\citet{liu2022learning}:} it 's a nice touch of <|unk|> severed \textbf{[15.23\%, 24.77\% too short]}\\

\noindent\textit{Original:} Vicis, a finalist for the 2016 GeekWire Awards in the Hardware/Gadget of the Year category, will use the fresh cash for production and commercialization of its ZERO1 helmet for the upcoming football season. \\
\textit{GPT3:} Vicis, a finalist for the 2016 GeekWire Awards in the Hardware/Gadget of the Year category, will use the fresh cash for production and commercialization of its ZERO1 helmet. \textbf{[83.98\%, 23.98\% too long]}\\
\textit{\secondphasemodelname{}:} Vicis, a finalist for the 2016 GeekWire Awards, will use the fresh cash for production and commercialization. \textbf{[52.91\%, in range]}\\
\textit{\citet{ghalandari2022efficient}:} Vicis, a finalist for 2016 GeekWire Awards/Gadget will use fresh cash for its ZERO1 helmet. \textbf{[44.17\%, in range]}\\
\textit{\citet{liu2022learning}:} a finalist for the of the year helmet the season \textbf{[23.3\%, 16.7\% too short]}\\

\noindent\textit{Original:} The cloud-based system has to wait for a connection to download updates to apps and content, but the content also has to be stored locally for uninterrupted playback. \\
\textit{GPT3:} The cloud-based system has to wait for a connection to download updates to apps and content, but the content also has to be stored locally for uninterrupted playback. \textbf{[100.0\%, 40.0\% too long]}\\
\textit{\secondphasemodelname{}:} The cloud-based system has to wait for a connection to download updates to apps and content. \textbf{[55.42\%, in range]}\\
\textit{\citet{ghalandari2022efficient}:} The cloud-based system has to wait for a connection to download updates to apps. \textbf{[48.19\%, in range]}\\
\textit{\citet{liu2022learning}:} the <|unk|> system has to wait for a download content stored locally \textbf{[40.96\%, in range]}\\

\noindent\textit{Original:} It signals toughness and finality, yet its meaning is elusive. \\
\textit{GPT3:} The article does not provide a clear explanation of what the phrase "yet its meaning is elusive" means. \textbf{[166.13\%, 106.13\% too long]}\\
\textit{\secondphasemodelname{}:} It signals toughness and finality. \textbf{[54.84\%, in range]}\\
\textit{\citet{ghalandari2022efficient}:} It signals toughness and finality \textbf{[53.23\%, in range]}\\
\textit{\citet{liu2022learning}:} it toughness its meaning is elusive \textbf{[56.45\%, in range]}\\

\noindent\subsection{Random sample of generations used in the 20-40\% human evaluation}
\textit{Original:} Viggo left South America aged 11, when his parents divorced and he moved to upstate New York with his mother and siblings. \\
\textit{GPT3:} Viggo left South America when he was 11. \textbf{[32.79\%, in range]}\\
\textit{\secondphasemodelname{}:} Viggo left South America aged 11, when his parents. \textbf{[41.8\%, 1.8\% too long]}\\
\textit{\citet{ghalandari2022efficient}:} Viggo left South America aged 11 parents divorced. \textbf{[40.98\%, 0.98\% too long]}\\
\textit{\citet{liu2022learning}:} parents moved to upstate new york \textbf{[27.05\%, in range]}\\

\noindent\textit{Original:} And Yahoo appears to have had another money-losing fumble on its hands with the free livestream of the NFL’s Oct. 25 Bills-Jaguars game from London. \\
\textit{GPT3:} Yahoo lost money streaming the NFL game from London. \textbf{[35.14\%, in range]}\\
\textit{\secondphasemodelname{}:} Yahoo appears to have had another money-losing fumble. \textbf{[36.49\%, in range]}\\
\textit{\citet{ghalandari2022efficient}:} And Yahoo had another money-losing fumbles Bills-Jaguars \textbf{[37.84\%, in range]}\\
\textit{\citet{liu2022learning}:} yahoo to have hands the of the nfl \textbf{[22.97\%, in range]}\\

\noindent\textit{Original:} Suárez almost levelled with his latest sweeping free-kick from distance but the clearer chances came to Hull against a ponderous Liverpool defence. \\
\textit{GPT3:} Hull scored against Liverpool. \textbf{[20.41\%, in range]}\\
\textit{\secondphasemodelname{}:} Suarez almost levelled with his latest free-kick. \textbf{[33.33\%, in range]}\\
\textit{\citet{ghalandari2022efficient}:} Suárez almost levelled with free-kick Hull Liverpool \textbf{[35.37\%, in range]}\\
\textit{\citet{liu2022learning}:} distance chances came to hull liverpool \textbf{[26.53\%, in range]}\\

\noindent\textit{Original:} Specifically, the company agreed to limit pre-payments, to provide accurate estimates of charges, and to disclose details of financing agreements. \\
\textit{GPT3:} The company agreed to limit pre-payments, to provide accurate estimates of charges, and to disclose details of financing agreements. \textbf{[90.41\%, 50.41\% too long]}\\
\textit{\secondphasemodelname{}:} The company agreed to limit pre-payments and to provide accurate estimates. \textbf{[51.37\%, 11.37\% too long]}\\
\textit{\citet{ghalandari2022efficient}:} Specifically company agreed to limit pre-payments. \textbf{[34.25\%, in range]}\\
\textit{\citet{liu2022learning}:} company to provide accurate disclose financing \textbf{[31.51\%, in range]}\\

\noindent\textit{Original:} "There was a family in the car, they got out before it hit. \\
\textit{GPT3:} There was a family in the car, they got out before it hit. \textbf{[98.31\%, 58.31\% too long]}\\
\textit{\secondphasemodelname{}:} "There was a family." \textbf{[35.59\%, in range]}\\
\textit{\citet{ghalandari2022efficient}:} "There was family car \textbf{[35.59\%, in range]}\\
\textit{\citet{liu2022learning}:} there was a it \textbf{[23.73\%, in range]}\\

\noindent\textit{Original:} As for the coming revolution, I can’t say for certain when that will happen. \\
\textit{GPT3:} I don't know when the revolution will happen. \textbf{[59.21\%, 19.21\% too long]}\\
\textit{\secondphasemodelname{}:} I can't say for sure. \textbf{[27.63\%, in range]}\\
\textit{\citet{ghalandari2022efficient}:} As for the coming revolution \textbf{[36.84\%, in range]}\\
\textit{\citet{liu2022learning}:} the revolution i can certain \textbf{[36.84\%, in range]}\\

\noindent\textit{Original:} In one New Jersey county, the Red Cross was AWOL, says Lt. Matthew Tiedemann of the Bergen County Office of Emergency Management. \\
\textit{GPT3:} The Red Cross was not present in one New Jersey county. \textbf{[42.64\%, 2.64\% too long]}\\
\textit{\secondphasemodelname{}:} In one New Jersey county, the Red Cross was AWOL. \textbf{[37.98\%, in range]}\\
\textit{\citet{ghalandari2022efficient}:} In New Jersey Red Cross was AWOL \textbf{[24.81\%, in range]}\\
\textit{\citet{liu2022learning}:} new jersey county red cross awol lt. \textbf{[27.91\%, in range]}\\

\noindent\textit{Original:} Furthermore, though some might argue that the purpose of food is merely to remind us of our own mortality, I would argue that the best cuisine can be an art form. \\
\textit{GPT3:} I think that food can be an art form. \textbf{[22.84\%, in range]}\\
\textit{\secondphasemodelname{}:} Furthermore, I would argue that the best cuisine can be an art form. \textbf{[41.98\%, 1.98\% too long]}\\
\textit{\citet{ghalandari2022efficient}:} Furthermore of food mortality the best cuisine art form. \textbf{[34.57\%, in range]}\\
\textit{\citet{liu2022learning}:} our mortality the best cuisine can be an art \textbf{[27.16\%, in range]}\\

\noindent\textit{Original:} Without proper conservation facilities nearby, the artifacts Batchvarov and his colleagues discovered had to be reburied underwater. \\
\textit{GPT3:} The artifacts Batchvarov and his colleagues discovered had to be reburied underwater. \textbf{[64.39\%, 24.39\% too long]}\\
\textit{\secondphasemodelname{}:} The artifacts had to be reburied. \textbf{[25.0\%, in range]}\\
\textit{\citet{ghalandari2022efficient}:} Without artifacts Batchvarov reburied underwater \textbf{[36.36\%, in range]}\\
\textit{\citet{liu2022learning}:} artifacts to be reburied underwater \textbf{[26.52\%, in range]}\\

\noindent\textit{Original:} Often people are desperate or ridiculous rather than heroic, so I do what needs to be done.” \\
\textit{GPT3:} The protagonist does what needs to be done. \textbf{[46.74\%, 6.74\% too long]}\\
\textit{\secondphasemodelname{}:} I do what needs to be done. \textbf{[29.35\%, in range]}\\
\textit{\citet{ghalandari2022efficient}:} Often people are desperate heroic. \textbf{[36.96\%, in range]}\\
\textit{\citet{liu2022learning}:} do what to be done \textbf{[19.57\%, 0.43\% too short]}\\

\noindent\textit{Original:} Liam’s feet make a guest appearance in the picture, as does his Great Dane Watson who he bought with ex-girlfriend Sophia Smith can be seen by his settee. \\
\textit{GPT3:} There are pictures of Liam and his dog Watson in the article. \textbf{[39.61\%, in range]}\\
\textit{\secondphasemodelname{}:} Liam's feet make a guest appearance in the picture. \textbf{[33.12\%, in range]}\\
\textit{\citet{ghalandari2022efficient}:} Liam's feet make guest appearance in picture Dane Watsongirlfriend. \textbf{[43.51\%, 3.51\% too long]}\\
\textit{\citet{liu2022learning}:} dane watson who bought ex-girlfriend sophia smith can be \textbf{[36.36\%, in range]}\\

\noindent\textit{Original:} I like this little gallery, the Small Weston Room, where the paintings go right up to the ceiling. \\
\textit{GPT3:} I like this little gallery. \textbf{[27.55\%, in range]}\\
\textit{\secondphasemodelname{}:} I like this little gallery. \textbf{[27.55\%, in range]}\\
\textit{\citet{ghalandari2022efficient}:} I like gallery Small Weston Room paintings \textbf{[42.86\%, 2.86\% too long]}\\
\textit{\citet{liu2022learning}:} go right to the ceiling \textbf{[23.47\%, in range]}\\

\noindent\textit{Original:} After a stint as a lobbyist, during which he helped establish two think tanks affiliated with James Dobson’s hugely influential Focus on the Family, Parker became a confidant of Roy Moore, then a county judge. \\
\textit{GPT3:} Parker became a confidant of Roy Moore and helped him win election as a county judge. \textbf{[40.67\%, 0.67\% too long]}\\
\textit{\secondphasemodelname{}:} Parker became a confidant of Roy Moore, then a county judge, during a stint. \textbf{[36.36\%, in range]}\\
\textit{\citet{ghalandari2022efficient}:} After as lobbyist Dobson, Parker became a confidant of Roy Moore. \textbf{[31.1\%, in range]}\\
\textit{\citet{liu2022learning}:} a a lobbyist dobson focus the family parker became a \textbf{[24.88\%, in range]}\\

\noindent\textit{Original:} Germany will be the favourites but it is wide open between the other three teams - Poland and Ukraine are a similar standard to Romania and Hungary, who we came up against in qualifying. \\
\textit{GPT3:} Germany is the favourite to win the World Cup, but it's still wide open. \textbf{[38.71\%, in range]}\\
\textit{\secondphasemodelname{}:} Germany is the favourite, but it is wide open between the other three teams. \textbf{[40.86\%, 0.86\% too long]}\\
\textit{\citet{ghalandari2022efficient}:} Germany will be the favourites but between teams Poland Ukraine. \textbf{[34.41\%, in range]}\\
\textit{\citet{liu2022learning}:} teams poland and ukraine standard romania hungary we came qualifying \textbf{[36.56\%, in range]}\\

\noindent\textit{Original:} It is the latest mass killing attributed to the extremists, activists said, with the violence coming during a 24-hour period. \\
\textit{GPT3:} Extremists have killed more people in 24 hours than ever before. \textbf{[51.2\%, 11.2\% too long]}\\
\textit{\secondphasemodelname{}:} It is the latest mass killing, activists said. \textbf{[36.8\%, in range]}\\
\textit{\citet{ghalandari2022efficient}:} It is latest mass killing attributed extremists \textbf{[37.6\%, in range]}\\
\textit{\citet{liu2022learning}:} latest mass killing attributed to extremists \textbf{[35.2\%, in range]}\\

\noindent\textit{Original:} The brutal honesty and reality of death speaks to the truth and tragedy of bad timing for these two characters.” \\
\textit{GPT3:} Death speaks to the truth and tragedy of bad timing in the characters' lives. \textbf{[68.75\%, 28.75\% too long]}\\
\textit{\secondphasemodelname{}:} The brutal honesty and reality of death. \textbf{[35.71\%, in range]}\\
\textit{\citet{ghalandari2022efficient}:} The brutal honesty death speaks characters. \textbf{[38.39\%, in range]}\\
\textit{\citet{liu2022learning}:} the reality the truth timing characters \textbf{[34.82\%, in range]}\\

\noindent\textit{Original:} It's a nice touch of confectionerial twinning, and broadly chimes with the optional theme of this year's show, 'Man Made', which judging by the parrots and skyscapes and severed heads and moths has not been taken up with widespread enthusiasm. \\
\textit{GPT3:} The exhibit has a theme. \textbf{[9.88\%, 10.12\% too short]}\\
\textit{\secondphasemodelname{}:} It's a nice touch of confectionery, and it's also a nice theme of this year's show. \textbf{[34.16\%, in range]}\\
\textit{\citet{ghalandari2022efficient}:} It's a nice touch of confectionerial twinning show 'Man Made. \textbf{[25.1\%, in range]}\\
\textit{\citet{liu2022learning}:} it 's a touch of the \textbf{[8.23\%, 11.77\% too short]}\\

\noindent\textit{Original:} Vicis, a finalist for the 2016 GeekWire Awards in the Hardware/Gadget of the Year category, will use the fresh cash for production and commercialization of its ZERO1 helmet for the upcoming football season. \\
\textit{GPT3:} Vicis will use the fresh cash to produce and commercialize its ZERO1 helmet. \textbf{[36.89\%, in range]}\\
\textit{\secondphasemodelname{}:} Vicis will use the fresh cash for production and commercialization. \textbf{[32.52\%, in range]}\\
\textit{\citet{ghalandari2022efficient}:} Vicis finalist GeekWire Awards/Gadget will use cash its ZERO1 helmet \textbf{[33.01\%, in range]}\\
\textit{\citet{liu2022learning}:} a finalist the of the year category helmet the season \textbf{[25.73\%, in range]}\\

\noindent\textit{Original:} The cloud-based system has to wait for a connection to download updates to apps and content, but the content also has to be stored locally for uninterrupted playback. \\
\textit{GPT3:} Cloud-based systems have to wait for updates, but the content can be stored locally for uninterrupted playback. \textbf{[66.87\%, 26.87\% too long]}\\
\textit{\secondphasemodelname{}:} The cloud-based system has to wait for a connection. \textbf{[31.33\%, in range]}\\
\textit{\citet{ghalandari2022efficient}:} The cloud-based system has to wait for download updates. \textbf{[33.73\%, in range]}\\
\textit{\citet{liu2022learning}:} <|unk|> has wait for a download content locally \textbf{[28.31\%, in range]}\\

\noindent\textit{Original:} It signals toughness and finality, yet its meaning is elusive. \\
\textit{GPT3:} The sign means something but we don't know what it is. \textbf{[87.1\%, 47.1\% too long]}\\
\textit{\secondphasemodelname{}:} It's a mystery. \textbf{[24.19\%, in range]}\\
\textit{\citet{ghalandari2022efficient}:} It signals toughness finality \textbf{[46.77\%, 6.77\% too long]}\\
\textit{\citet{liu2022learning}:} signals toughness elusive \textbf{[40.32\%, 0.32\% too long]}\\

\end{document}